\documentclass[10pt,journal,compsoc]{IEEEtran}
\usepackage[caption=false,font=footnotesize,labelfont=sf,textfont=sf]{subfig}
\usepackage{algorithmic}
\usepackage[utf8]{inputenc}
\usepackage{graphicx}
\usepackage{calc}
\usepackage[normalem]{ulem}
\usepackage{tikz,pgfplots}
\usepackage{grffile}
\usepackage{amsmath}
\usepackage{amsthm}
\usepackage{breqn}
\usepackage{gensymb}
\usepackage{url}
\usepackage[bookmarks=true,bookmarksnumbered=true,
pdfpagemode={UseOutlines},plainpages=false,pdfpagelabels=true,
pdftitle={Increasing Impact by Mechanical Resonance for Teleoperated Hammering},
pdfsubject={Telemanipulation with series elastic actuators},
pdfauthor={Manuel Aiple, Jan Smisek, and Andre Schiele},
pdfkeywords={Biomechanics, force feedback, human performance, telemanipulation}]{hyperref}
\usepackage{array}
\usepackage{xspace}
\usepackage{verbatim}
\usepackage{booktabs}
\usepackage{blindtext}
\usepackage[flushleft]{threeparttable}
\usepackage[shortcuts]{extdash}
\usepackage{fancyref}
\usepackage{colortbl}
\usepackage{placeins}
\usepackage{siunitx}
\usepackage{todo}

\pgfplotsset{compat=newest}
\usetikzlibrary{plotmarks}
\usetikzlibrary{arrows.meta}
\usepgfplotslibrary{patchplots}
\pgfplotsset{plot coordinates/math parser=false}
\newlength\figureheight
\newlength\figurewidth

\usepackage{mathtools}

\DeclarePairedDelimiter\abs{\lvert}{\rvert}%
\DeclarePairedDelimiter\norm{\lVert}{\rVert}%

\usetikzlibrary{shapes,arrows,calc,intersections,positioning,fit}

\newcommand{\plotheightA}{.18\textheight}
\newcommand{\plotheightB}{.35\textheight}

\definecolor{printer1}{cmyk}{0.10,0.90,0.80,0.0}%
\definecolor{printer2}{cmyk}{0.80,0.30,0.0,0.0}%
\definecolor{printer3}{cmyk}{0.70,0.0,0.80,0.0}%
\definecolor{printer4}{cmyk}{0.40,0.65,0.0,0.0}%
\definecolor{printer5}{cmyk}{0.0,0.50,1.00,0.0}%
\definecolor{printer6}{cmyk}{0.0,0.0,0.80,0.0}%
\definecolor{printer7}{cmyk}{0.35,0.60,0.80,0.0}%
\definecolor{printer8}{cmyk}{0.0,0.50,0.0,0.0}%
\definecolor{printer9}{cmyk}{0.0,0.0,0.0,0.40}%
\definecolor{visible1}{cmyk}{0.35,0.07,0.0,0.0}%
\definecolor{visible2}{cmyk}{0.90,0.30,0.0,0.0}%
\definecolor{visible3}{cmyk}{0.30,0.0,0.45,0.0}%
\definecolor{visible4}{cmyk}{0.80,0.0,1.00,0.0}%
\definecolor{visible5}{cmyk}{0.0,0.40,0.25,0.0}%
\definecolor{visible6}{cmyk}{0.10,0.90,0.80,0.0}%
\definecolor{visible7}{cmyk}{0.0,0.25,0.50,0.0}%
\definecolor{visible8}{cmyk}{0.0,0.50,1.00,0.0}%
\definecolor{visible9}{cmyk}{0.20,0.25,0.0,0.0}%
\definecolor{visible10}{cmyk}{0.60,0.70,0.0,0.0}%
\definecolor{visible11}{cmyk}{0.0,0.0,0.40,0.0}%
\definecolor{visible12}{cmyk}{0.23,0.73,0.98,0.12}%
\definecolor{dark1}{cmyk}{0.90,0.0,0.55,0.0}%
\definecolor{dark2}{cmyk}{0.15,0.60,1.00,0.0}%
\definecolor{dark3}{cmyk}{0.55,0.45,0.0,0.0}%
\definecolor{dark4}{cmyk}{0.05,0.85,0.05,0.0}%
\definecolor{dark5}{cmyk}{0.60,0.10,1.00,0.0}%
\definecolor{dark6}{cmyk}{0.10,0.30,1.00,0.0}%
\definecolor{dark7}{cmyk}{0.35,0.45,0.90,0.0}%
\definecolor{dark8}{cmyk}{0.0,0.0,0.0,0.60}%
\definecolor{gray1}{gray}{0.0}%
\definecolor{gray2}{gray}{0.4}%
\definecolor{gray3}{gray}{0.8}%
\definecolor{matlab1}{rgb}{0.00000,0.44700,0.74100}%
\definecolor{matlab2}{rgb}{0.85000,0.32500,0.09800}%
\definecolor{matlab3}{rgb}{0.92900,0.69400,0.12500}%
\definecolor{matlab4}{rgb}{0.49400,0.18400,0.55600}%
\definecolor{matlab5}{rgb}{0.46600,0.67400,0.18800}%
\definecolor{matlab6}{rgb}{0.3,0.3,0.3}%
\definecolor{matlab7}{rgb}{0.30100,0.74500,0.93300}%
\definecolor{matlab8}{rgb}{0.63500,0.07800,0.18400}%
\definecolor{matlab9}{rgb}{0.7,0.7,0.7}%

\newlength{\dotSymRad}
\newlength{\hammerlength}
\newlength{\estoplength}
\newcommand{\tikzAngleOfLine}{\tikz@AngleOfLine}
\def\tikz@AngleOfLine(#1)(#2)#3{%
	\pgfmathanglebetweenpoints{%
		\pgfpointanchor{#1}{center}}{%
		\pgfpointanchor{#2}{center}}
	\pgfmathsetmacro{#3}{\pgfmathresult}%
}

\newcommand{\junction}[2]{%
	\draw[fill] let \p1 = #1,\p2 = #2 in (\x1,\y2) circle (1pt);
}

\newlength{\gndsymwidth}
\newlength{\gndsymheight}
\newlength{\lindampersymheight}
\newlength{\lindampersymwidth}
\newlength{\linspringsymheight}
\newlength{\linspringsymwidth}
\newlength{\masssymrad}
\newcommand{\gndLeft}[2]{%
	\coordinate (gnd#2) at #1;
	\draw #1 -- ($#1-(.7\gndsymheight,0cm)$);
	\draw ($#1-(.7\gndsymheight,-.5\gndsymwidth)$) -- ($#1-(.7\gndsymheight,.5\gndsymwidth)$);
	\draw ($#1-(\gndsymheight,-.35\gndsymwidth)$) -- ($#1-(.7\gndsymheight,-.45\gndsymwidth)$);
	\draw ($#1-(\gndsymheight,-.25\gndsymwidth)$) -- ($#1-(.7\gndsymheight,-.35\gndsymwidth)$);
	\draw ($#1-(\gndsymheight,-.15\gndsymwidth)$) -- ($#1-(.7\gndsymheight,-.25\gndsymwidth)$);
	\draw ($#1-(\gndsymheight,-.05\gndsymwidth)$) -- ($#1-(.7\gndsymheight,-.15\gndsymwidth)$);
	\draw ($#1-(\gndsymheight,.05\gndsymwidth)$) -- ($#1-(.7\gndsymheight,-.05\gndsymwidth)$);
	\draw ($#1-(\gndsymheight,.15\gndsymwidth)$) -- ($#1-(.7\gndsymheight,.05\gndsymwidth)$);
	\draw ($#1-(\gndsymheight,.25\gndsymwidth)$) -- ($#1-(.7\gndsymheight,.15\gndsymwidth)$);
	\draw ($#1-(\gndsymheight,.35\gndsymwidth)$) -- ($#1-(.7\gndsymheight,.25\gndsymwidth)$);
	\draw ($#1-(\gndsymheight,.45\gndsymwidth)$) -- ($#1-(.7\gndsymheight,.35\gndsymwidth)$);
}

\newcommand{\lindamperLeft}[4]{%
	\coordinate (lindamp#2) at #1;
	\coordinate (lindampeast#2) at ($(lindamp#2)+(.5\lindampersymwidth,0cm)$);
	\coordinate (lindampwest#2) at ($(lindamp#2)+(-.5\lindampersymwidth,0cm)$);
	\draw ($(lindamp#2)+(.3\lindampersymwidth,.5\lindampersymheight)$) -- ($(lindamp#2)+(-.3\lindampersymwidth,.5\lindampersymheight)$);
	\draw ($(lindamp#2)+(.3\lindampersymwidth,-.5\lindampersymheight)$) -- ($(lindamp#2)+(-.3\lindampersymwidth,-.5\lindampersymheight)$);
	\draw ($(lindamp#2)+(.3\lindampersymwidth,.5\lindampersymheight)$) -- ($(lindamp#2)+(.3\lindampersymwidth,-.5\lindampersymheight)$);
	\draw ($(lindamp#2)+(-.1\lindampersymwidth,+.5\lindampersymheight)$) -- ($(lindamp#2)+(-.1\lindampersymwidth,-.5\lindampersymheight)$);
	\draw ($(lindamp#2)+(-.5\lindampersymwidth,0cm)$) -- ($(lindamp#2)+(-.1\lindampersymwidth,0cm)$);
	\draw ($(lindamp#2)+(.3\lindampersymwidth,0cm)$) -- ($(lindamp#2)+(.5\lindampersymwidth,0cm)$);
	\node (lindamplabel#2) [node distance=.5\lindampersymheight,#4=of lindamp#2]{#3};
}
\newcommand{\lindamperRight}[4]{%
	\coordinate (lindamp#2) at #1;
	\coordinate (lindampeast#2) at ($(lindamp#2)+(.5\lindampersymwidth,0cm)$);
	\coordinate (lindampwest#2) at ($(lindamp#2)+(-.5\lindampersymwidth,0cm)$);
	\draw ($(lindamp#2)+(.3\lindampersymwidth,.5\lindampersymheight)$) -- ($(lindamp#2)+(-.3\lindampersymwidth,.5\lindampersymheight)$);
	\draw ($(lindamp#2)+(.3\lindampersymwidth,-.5\lindampersymheight)$) -- ($(lindamp#2)+(-.3\lindampersymwidth,-.5\lindampersymheight)$);
	\draw ($(lindamp#2)+(-.3\lindampersymwidth,.5\lindampersymheight)$) -- ($(lindamp#2)+(-.3\lindampersymwidth,-.5\lindampersymheight)$);
	\draw ($(lindamp#2)+(.1\lindampersymwidth,+.5\lindampersymheight)$) -- ($(lindamp#2)+(.1\lindampersymwidth,-.5\lindampersymheight)$);
	\draw ($(lindamp#2)+(-.5\lindampersymwidth,0cm)$) -- ($(lindamp#2)+(-.3\lindampersymwidth,0cm)$);
	\draw ($(lindamp#2)+(.1\lindampersymwidth,0cm)$) -- ($(lindamp#2)+(.5\lindampersymwidth,0cm)$);
	\node (lindamplabel#2) [node distance=.5\lindampersymheight,#4=of lindamp#2]{#3};
}
\newcommand{\linspring}[4]{%
	\coordinate (linspring#2) at #1;
	\coordinate (linspringeast#2) at ($(linspring#2)+(.5\linspringsymwidth,0cm)$);
	\coordinate (linspringwest#2) at ($(linspring#2)+(-.5\linspringsymwidth,0cm)$);
	\draw ($(linspring#2)+(-.5\linspringsymwidth,0cm)$) -- ($(linspring#2)+(-.3\linspringsymwidth,0cm)$);
	\draw ($(linspring#2)+(.5\linspringsymwidth,0cm)$) -- ($(linspring#2)+(.3\linspringsymwidth,0cm)$);
	\draw ($(linspring#2)+(-.3\linspringsymwidth,0cm)$) -- ($(linspring#2)+(-.25\linspringsymwidth,+.5\linspringsymheight)$);
	\draw ($(linspring#2)+(.3\linspringsymwidth,0cm)$) -- ($(linspring#2)+(+.25\linspringsymwidth,-.5\linspringsymheight)$);
	\draw ($(linspring#2)+(-.25\linspringsymwidth,+.5\linspringsymheight)$) -- ($(linspring#2)+(-.15\linspringsymwidth,-.5\linspringsymheight)$);
	\draw ($(linspring#2)+(-.15\linspringsymwidth,-.5\linspringsymheight)$) -- ($(linspring#2)+(-.05\linspringsymwidth,.5\linspringsymheight)$);
	\draw ($(linspring#2)+(-.05\linspringsymwidth,.5\linspringsymheight)$) -- ($(linspring#2)+(.05\linspringsymwidth,-.5\linspringsymheight)$);
	\draw ($(linspring#2)+(.05\linspringsymwidth,-.5\linspringsymheight)$) -- ($(linspring#2)+(.15\linspringsymwidth,+.5\linspringsymheight)$);
	\draw ($(linspring#2)+(.15\linspringsymwidth, +.5\linspringsymheight)$) -- ($(linspring#2)+(.25\linspringsymwidth,-.5\linspringsymheight)$);
	\node (linspringlabel#2) [node distance=.5\linspringsymheight,#4=of linspring#2]{#3};
}

\newcommand{\massSimple}[3]{%
	\draw #1 circle (\masssymrad);
	\node[minimum size=2\masssymrad,circle] (mass#2) at #1 {#3};
}
\newcommand{\forcevector}[5]{%
	\coordinate (forceP1#3) at #1;
	\coordinate (forceP2#3) at #2;
	\draw[>=latex,->] (forceP1#3) -- (forceP2#3);
	\node (forcelabel#3) [node distance=0.2em,#5=of $(forceP1#3)!0.5!(forceP2#3)$]{#4};
}
\newcommand{\velocityvector}[5]{%
	\coordinate (velocityP1#3) at #1;
	\coordinate (velocityP2#3) at #2;
	\draw[>=latex,->] (velocityP1#3) -- (velocityP2#3);
	\node (velocitylabel#3) [node distance=0.2em,#5=of $(velocityP1#3)!0.5!(velocityP2#3)$]{#4};
}

\newcommand{\kstiffness}{\ensuremath{k}\xspace}

\newcommand{\Kvar}{\ensuremath{K}\xspace}
\newcommand{\mmass}{\ensuremath{m}\xspace}
\newcommand{\bdamping}{\ensuremath{b}\xspace}
\newcommand{\Fforce}{\ensuremath{F}\xspace}
\newcommand{\ffrequency}{\ensuremath{f}\xspace}
\newcommand{\fresonance}{\ensuremath{f_0}\xspace}

\newcommand{\Ggain}{\ensuremath{G}\xspace}

\newcommand{\Zh}{\ensuremath{Z_h}\xspace}
\newcommand{\Ze}{\ensuremath{Z_e}\xspace}
\newcommand{\Zto}{\ensuremath{Z_{to}}\xspace}
\newcommand{\fh}{\ensuremath{f_h}\xspace}
\newcommand{\fhext}{\ensuremath{f_{\tilde{h}}}\xspace}
\newcommand{\fe}{\ensuremath{f_e}\xspace}
\newcommand{\feext}{\ensuremath{f_{\tilde{e}}}\xspace}

\newcommand{\fmc}{\ensuremath{f_{mc}}\xspace}

\newcommand{\fsc}{\ensuremath{f_{sc}}\xspace}

\newcommand{\mm}{\ensuremath{m_m}\xspace}

\newcommand{\vm}{\ensuremath{\dot{x}_m}\xspace}
\newcommand{\vs}{\ensuremath{\dot{x}_s}\xspace}

\newcommand{\xdelta}{\ensuremath{\Delta x}\xspace}

\newcommand{\vin}{\ensuremath{v_{in}}\xspace}

\newcommand{\vflx}{\ensuremath{v_{flx}}\xspace}
\newcommand{\vrig}{\ensuremath{v_{rig}}\xspace}
\newcommand{\Vin}{\ensuremath{V_{in}}\xspace}

\newcommand{\Vflx}{\ensuremath{V_{flx}}\xspace}
\newcommand{\Vrig}{\ensuremath{V_{rig}}\xspace}

\newcommand{\slap}{\ensuremath{s}\xspace}

\newcommand{\ttime}{\ensuremath{t}\xspace}
\newcommand{\Tdelaylap}{\ensuremath{e^{-s T}}\xspace}
\newcommand{\Tdelay}{\ensuremath{T}\xspace}

\newcommand{\Cm}{\ensuremath{C_m}\xspace}
\newcommand{\Cs}{\ensuremath{C_s}\xspace}
\newcommand{\Cone}{\ensuremath{C_1}\xspace}
\newcommand{\Ctwo}{\ensuremath{C_2}\xspace}
\newcommand{\Cthree}{\ensuremath{C_3}\xspace}
\newcommand{\Cfour}{\ensuremath{C_4}\xspace}

\newcommand{\Zs}{\ensuremath{Z_s}\xspace}
\newcommand{\Zsinv}{\ensuremath{Z_s^{-1}}\xspace}
\newcommand{\Zm}{\ensuremath{Z_m}\xspace}
\newcommand{\Zminv}{\ensuremath{Z_m^{-1}}\xspace}

\newcommand{\pvar}{\ensuremath{p}\xspace}
\newcommand{\nvar}{\ensuremath{n}\xspace}
\newcommand{\alphavar}{\ensuremath{\alpha}\xspace}
\newcommand{\chisquarevar}{\ensuremath{\chi^2}\xspace}

\newcommand{\bvalue}[1]{\ensuremath{b_{#1}}\xspace}
\newcommand{\FFcond}{FF}
\newcommand{\NVcond}{NV}
\newcommand{\NFcond}{NF}
\newcommand{\DLcond}{DL}
\newcommand{\Ekinetic}{\ensuremath{E_k}\xspace}
\newcommand{\Espring}{\ensuremath{E_p}\xspace}
\newcommand{\vvelocity}{\ensuremath{v}\xspace}

\newcommand{\Tperiod}{\ensuremath{T}\xspace}

\newcommand{\Hrigid}{\ensuremath{H_{rigid}}\xspace}
\newcommand{\Hflexible}{\ensuremath{H_{flexible}}\xspace}

\newcommand{\Hmax}{\ensuremath{H_{max}}\xspace}

\newcommand{\Tresonance}{\ensuremath{T_0}\xspace}
\newcommand{\Topt}{\ensuremath{T^{*}}\xspace}
\newcommand{\Ehatrigid}{\ensuremath{\hat{E}_{rig}}\xspace}
\newcommand{\Ehatflex}{\ensuremath{\hat{E}_{flx}}\xspace}
\newcommand{\vhatrigid}{\ensuremath{\hat{v}_{rig}}\xspace}
\newcommand{\vhatflex}{\ensuremath{\hat{v}_{flx}}\xspace}

\newcommand{\vhatin}{\ensuremath{\hat{v}_{in}}\xspace}
\newcommand{\vhat}{\ensuremath{\hat{v}}\xspace}
\newcommand{\CI}[1][95]{\ensuremath{\text{CI}_{#1\%}}\xspace}
\newcommand{\jcomplex}{\ensuremath{j}\xspace}
\newcommand{\omegarotvel}{\ensuremath{\omega}\xspace}

\newcommand{\sigmastdest}{\ensuremath{\overline{\sigma}}\xspace}
\newcommand{\madfunc}{\ensuremath{\text{MAD}}\xspace}
\newcommand{\ExpOne}{Experiment~1\xspace}
\newcommand{\ExpTwo}{Experiment~2\xspace}

\colorlet{colorFF}{matlab1}
\colorlet{colorNV}{matlab2}
\colorlet{colorNF}{matlab3}
\colorlet{colorDL}{matlab4}
\colorlet{color3p0Hz}{matlab5}
\colorlet{color4p8Hz}{matlab6}
\colorlet{color6p9Hz}{matlab7}
\colorlet{color9p9Hz}{matlab8}
\colorlet{colorRig}{matlab9}
\colorlet{colorFlexible}{visible4}
\colorlet{colorRigid}{visible2}

\begin{document}
\title{Increasing Impact by Mechanical Resonance for Teleoperated Hammering}
\author{Manuel Aiple, Jan Smisek, and André Schiele%
\thanks{M. Aiple and A. Schiele are with the BioMechanical Engineering Department, Faculty of Mechanical, Maritime and Materials Engineering, Delft University of Technology, Netherlands.
J. Smisek is with the Telerobotics and Haptics Lab, European Space Agency, Noordwijk, Netherlands and the Department of Control and Simulation, Faculty of Aerospace Engineering, Delft University of Technology, Netherlands.\protect\\%
E-mail: \texttt{m.aiple@tudelft.nl}%
}}

\IEEEtitleabstractindextext{%
\begin{abstract}
Series elastic actuators (SEAs) are interesting for usage in harsh environments as they are more robust than rigid actuators. This paper shows how SEAs can be used in teleoperation to increase output velocity in dynamic tasks.
A first experiment is presented that tested human ability to achieve higher hammerhead velocities with a flexible hammer than with a rigid hammer, and to evaluate the influence of the resonance frequency.
In this experiment, 13 participants executed a hammering task in direct manipulation using flexible hammers in four conditions with resonance frequencies of \SIrange{3.0}{9.9}{\Hz} and one condition with a rigid hammer.
Then, a second experiment is presented that tested the ability of 32 participants to reproduce the findings of the first experiment in teleoperated manipulation with different feedback conditions: with visual and force feedback, without visual feedback, without force feedback, and with a communication delay of \SI{40}{\ms}.
The results indicate that humans can exploit the mechanical resonance of a flexible system to at least double the output velocity without combined force and vision feedback.
This is an unexpected result, allowing the design of simpler and more robust teleoperators for dynamic tasks.
\end{abstract}%
}

\maketitle
\IEEEraisesectionheading{%
\section{Introduction}%
\label{sec:introduction}%
}

\IEEEPARstart{T}{eleoperation} is classically desired to be performed with a transparent system \cite{Niemeyer2008}. This is often realized by using mechanically stiff devices.
While this is advantageous for precision, it is an issue for dynamic tasks, where high impacts can be experienced during contact, such as in harsh environments in search and rescue missions~\cite{Trevelyan2016}, explosive ordnance disposal~\cite{Trevelyan2016}, nuclear research maintenance~\cite{Boessenkool2017}, teleoperated forestry~\cite{Milne}, or underwater robotics~\cite{Bruzzone2003}.
A stiff robot experiences high shock loads on impact, which can damage the robot's gears \cite{Pratt1995}.
Series elastic actuators (SEAs) promise higher mechanical robustness \cite{Pratt1995}, but their usage for precision \cite{Christiansson2007} and high-impact tasks in teleoperation still requires research.

One interesting feature of SEAs is the possibility to maximize power transfer from an operator controlling a handle device (``master device'' in teleoperation literature) to a tool device (``slave device''), if the mechanical resonance intrinsic to SEAs is well exploited during the movement.
By ``exploit'' we mean sense and utilize the system's resonance frequency and excite it to achieve maximum power transfer and maximum output velocity.
Until now, there have been only a few studies on maximum power transfer in SEAs, with the focus on designing fully automatic control laws~\cite{Braun2012a,Garabini2011,Haddadin2011}.

However, to the knowledge of the authors, the following three questions have received little attention in previous research.
1) Can humans sense and identify the resonance frequency of SEAs?
2) Within which resonance frequency/stiffness range can humans sense the resonance frequency and tune their motor actions to the timing required to exploit SEAs for maximum power transfer?
3) What type and quality of feedback should be given to humans to exploit SEAs for maximum power transfer?
Answering these open questions will enable the design of new teleoperation systems with variable stiffness tool actuators with the objective of allowing not only precise positioning, but also the performance of high-impact tasks.

\begin{figure}[!t]
    \centering
    {
	\footnotesize
    \input{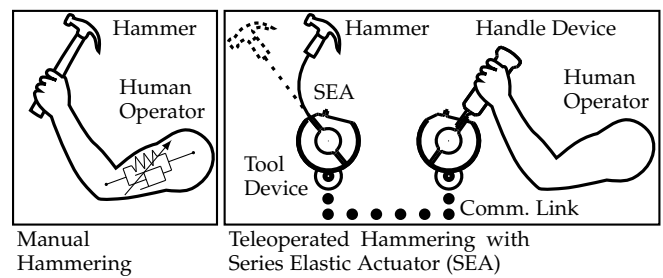}
    }
    \caption{This study is part of a research project aimed at developing methods to allow the execution of human real-life dynamic tasks, such as hammering (left) using a teleoperation system with an SEA tool (right).
    Mechanical compliance, which is inherent to human muscles, is achieved through a spring in the robotic tool.}
    \label{fig:conceptpicture}
\end{figure}

The aim of the present study was to explore these questions experimentally by having human operators execute a hammering task with a haptic teleoperated single degree of freedom hammering apparatus. To this end, two experiments were conceived.
The first was designed to investigate the first two questions above and to determine:
a) the influence of frequency/stiffness on human performance in exploiting SEAs for maximum power transfer,
b) the degree to which humans can achieve the optimal excitation timing (see section~\ref{sec:resonance-explanation} for a definition of the optimal timing), and
c) the practically achievable output velocity to input velocity amplification.
We will explain in the following section how a flexible system can achieve a higher power transfer than a rigid system.
The second experiment was designed to approach the third question by quantifying the dependence of the human manipulation of flexible tools concerning the availability of visual and force feedback, as well as their sensitivity to the presence of time delay.

In both experiments, hammering was selected as the experimental task, as the principles are generally known and the task itself can be simplified to one degree of freedom.
It might seem paradoxical to use an elastic hammer, which can less effectively transmit forces through the handle to the hammerhead.
However, the primary objective of hammering is to transform kinetic energy into mechanical work by the impact.
The effect of the force transmitted through the handle during the impact is negligible in comparison.
Consequently, typical hammers have longer handles to obtain high hammerhead velocities (while keeping the same hand velocity).
Furthermore, hammers in practice are held loosely at the moment of impact to reduce the shock load to the hands and arms.
The same principle is applied in power machines, such as pile drivers, jackhammers, and hammer drills, where the hammerhead is either running free during the impact or is only loosely coupled to the actuator.

We advocate that hammering with an elastic hammer could have a double benefit for teleoperated hammering:
1) an elastic hammer could achieve a higher impact velocity than a rigid hammer with the same velocity of the tool actuator and
2) the shock loads to the tool actuator could be reduced.
Such a system could provide a better performance and have a longer lifetime without requiring more powerful actuators or stronger mechanical structures.

\section{Increasing Impact with a Flexible Hammer by Mechanical Resonance}
\label{sec:resonance-explanation}
\subsection{Modeling of the Rigid and the Flexible Hammer}
The following explanations are illustrated with a linear mass-spring-damper model instead of a rotational mass-spring-damper model, as the equations are the same for linear variables and their rotational equivalents.
The transfer function $\Hrigid(\slap)$ of a mass $\mmass > 0$, exposed to viscous damping $\bdamping > 0$, as illustrated in Fig.~\ref{fig:rigid-tool-model}, is trivially described in the Laplace domain by 
\begin{dmath}
	\Hrigid(\slap) \hiderel{=} \frac{\Vrig(\slap)}{\Vin(\slap)} = 1
	\label{eq:Hrigid}
\end{dmath},
with $\Vin(\slap)$ and $\Vrig(\slap)$ the Laplace-transformation of $\vin(\ttime)$ and $\vrig(\ttime)$, which are the input velocity and the velocity of the rigid hammer.

The transfer function of a mass-spring-damper system with stiffness $\kstiffness > 0$ as modeled in Fig.~\ref{fig:flexible-tool-model} is obtained from the equality of forces acting on the mass \mmass as
\begin{dmath}
(\slap\,\mmass + \bdamping) \; \Vflx(\slap) = -\frac{1}{\slap}\,\kstiffness\;(\Vflx(\slap) - \Vin(\slap))
\end{dmath},
with $\Vflx(\slap)$ the Laplace-transformation of $\vflx(\ttime)$, which is the velocity of the flexible hammer,
resulting in
\begin{dmath}
	\Hflexible(\slap) \hiderel{=} \frac{\Vflx(\slap)}{\Vin(\slap)}
	= \frac{k}{s^2\,m + s\,b + k}
\label{eq:Hflexible}
\end{dmath}.

The resonance frequency \fresonance is the frequency \ffrequency for which the magnitude of $\Hflexible(\slap)$ is maximized.
This is obtained by solving the minimization problem of the denominator:
\begin{dmath}
	\min_{\slap}\; \norm{\slap^2\,\mmass + \slap\,\bdamping + \kstiffness}
	\label{eq:minproblem}
\end{dmath},
with the Laplace variable
\begin{math}
\slap = \jcomplex\,\omegarotvel
\end{math},
and the angular frequency
\begin{math}
\omegarotvel = 2\,\pi\,\ffrequency > 0
\end{math}.
Equation~\ref{eq:minproblem} is solved by the resonance frequency
\begin{dmath}
\fresonance = \frac{1}{2\,\pi}\;\sqrt{\frac{\kstiffness}{\mmass}-\frac{b^2}{2\,\mmass^2}}
\end{dmath}.
Because the term under the square-root has to be positive for \fresonance to have a real value, the criterion for resonance to occur is
\begin{dmath}\bdamping < \sqrt{2\,\kstiffness\,\mmass}\end{dmath}.
In the following, we assume that this condition is satisfied and that
\begin{math}\bdamping \ll \sqrt{2\,\kstiffness\,\mmass}\end{math}.
The equation for the resonance frequency can be then simplified to
\begin{dmath}
\fresonance \hiderel{\approx} \frac{1}{2\,\pi}\;\sqrt{\frac{\kstiffness}{\mmass}}
\label{eq:fresonance}
\end{dmath}.

\subsection{Exploiting Mechanical Resonance for Increased Output Velocity}
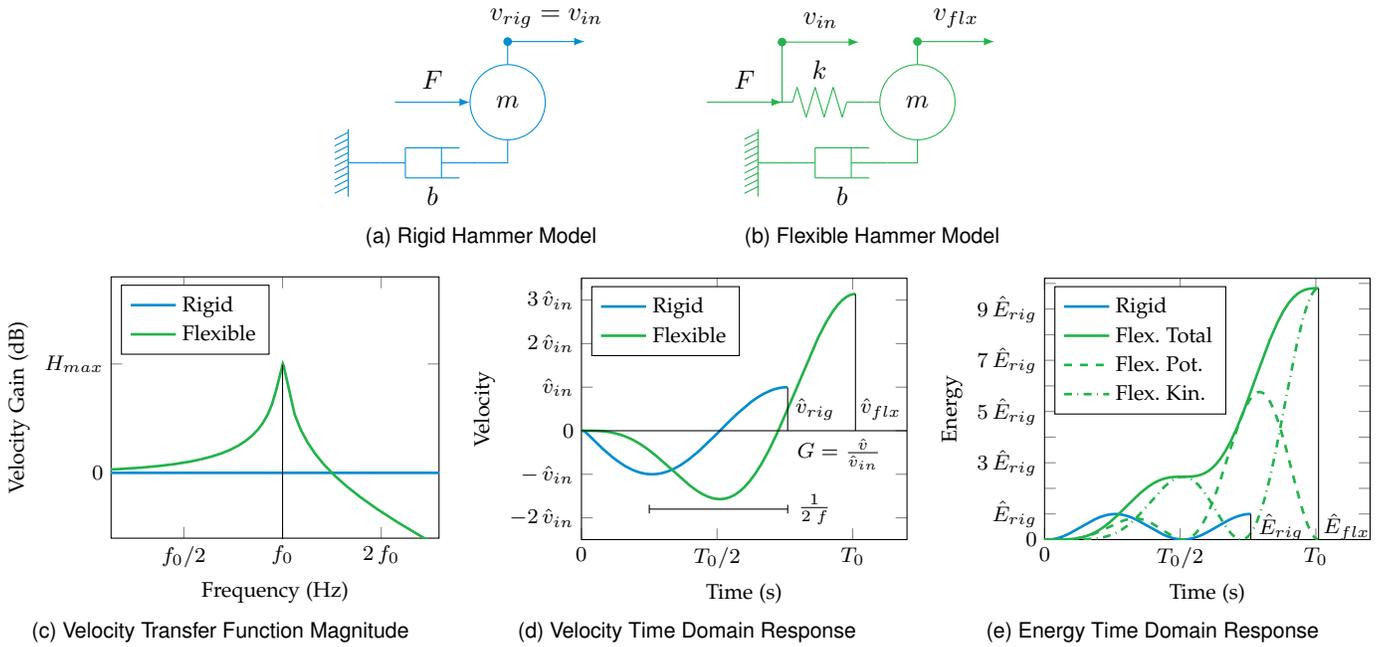
\begin{figure*}[!t]
	\centering
	\setlength{\figurewidth}{.25\textwidth}
	\setlength{\figureheight}{.14\textheight}
	\subfloat[Rigid Hammer Model]{\begin{tikzpicture}
\begin{scope}[every path/.style={draw=colorRigid}]
\coordinate (rigidref) at (0cm,0cm);
\draw[white] ($(rigidref)+(-1cm,-1.5cm)$) rectangle ($(rigidref)+(3.2cm,1cm)$);
\coordinate (finjunctionrig) at ($(rigidref)+(1cm,0cm)$);

\forcevector{(rigidref)}{(finjunctionrig)}{rig}{$\Fforce$}{above}
\massSimple{($(finjunctionrig)+(\masssymrad,0cm)$)}{rigid}{$\mmass$}
\coordinate (foutjunctionrig) at (massrigid.east);
\coordinate (foutrig) at ($(massrigid.east)+(1cm,0cm)$);
\coordinate (vrigidref) at ($(massrigid.north)+(0cm,0.3cm)$);
\draw[fill=colorRigid] (vrigidref) circle (0.05cm);
\coordinate (vinrig) at ($(vrigidref)+(-1.025cm,0cm)$);
\coordinate (voutrig) at ($(vrigidref)+(1.025cm,0cm)$);
\velocityvector{(vrigidref)}{(voutrig)}{flex}{$\vrig = \vin$}{above}
\lindamperRight{($(massrigid.south)+(-1cm,-0.3cm)$)}{rigid}{$\bdamping$}{below}
\gndLeft{($(lindampwestrigid)+(-0.1cm,0cm)$)}{chassisrigid}
\draw (lindampwestrigid) -- (gndchassisrigid);
\draw (lindampeastrigid) -| (massrigid.south);
\draw (massrigid.north) -- (vrigidref);
\end{scope}
\end{tikzpicture}\label{fig:rigid-tool-model}}
	\hspace{.03\textwidth}
	\subfloat[Flexible Hammer Model]{\begin{tikzpicture}
\begin{scope}[every path/.style={draw=colorFlexible}]
\coordinate (flexref) at (0cm,0cm); 
\draw[white] ($(flexref)+(-0.2cm,-1.5cm)$) rectangle ($(flexref)+(4.5cm,1cm)$);
\coordinate (finjunctionflex) at ($(flexref)+(1cm,0cm)$);
\forcevector{(flexref)}{(finjunctionflex)}{flex}{$\Fforce$}{above}
\linspring{($(finjunctionflex)+(0.5cm,0cm)$)}{flex}{$\kstiffness$}{above}
\coordinate (flexmassjunction) at ($(linspringeastflex)+(0.1cm,0cm)$);
\massSimple{($(flexmassjunction)+(\masssymrad+0.1cm,0cm)$)}{flexible}{$\mmass$}
\coordinate (vflexinref) at ($(finjunctionflex)+(0cm,\masssymrad+0.3cm)$);
\draw[fill=colorFlexible] (vflexinref) circle (0.05cm);
\coordinate (vflexoutref) at ($(massflexible.north)+(0cm,0.3cm)$);
\draw[fill=colorFlexible] (vflexoutref) circle (0.05cm);
\coordinate (vinflex) at ($(vflexinref)+(1.025cm,0cm)$);
\coordinate (voutflex) at ($(vflexoutref)+(1.025cm,0cm)$);
\velocityvector{(vflexinref)}{(vinflex)}{rig}{$\vin$}{above}
\velocityvector{(vflexoutref)}{(voutflex)}{rig}{$\vflx$}{above}
\draw (finjunctionflex) -- (vflexinref);
\draw (massflexible.north) -- (vflexoutref);
\draw (finjunctionflex) |- (linspringwestflex);
\draw (flexmassjunction) |- (linspringeastflex);
\draw (flexmassjunction) -- (massflexible.west);
\lindamperRight{($(massflexible.south)+(-1cm,-0.3cm)$)}{flexible}{$\bdamping$}{below}
\gndLeft{($(lindampwestflexible)+(-0.1cm,0cm)$)}{chassisflexible}
\draw (lindampwestflexible) -- (gndchassisflexible);
\draw (lindampeastflexible) -| (massflexible.south);
\end{scope}
\end{tikzpicture}\label{fig:flexible-tool-model}}\\
	{\footnotesize
	\subfloat[Velocity Transfer Function Magnitude]{
%
%
\colorlet{mycolor1}{colorRigid}
\colorlet{mycolor2}{colorFlexible}
\begin{tikzpicture}

\begin{axis}[%
width=0.958\figurewidth,
height=\figureheight,
at={(0\figurewidth,0\figureheight)},
scale only axis,
xmode=log,
xmin=0.3,
xmax=3,
xtick={0.5,1,2},
xticklabels={
{$f_0/2$},
{$f_0$},
{$2\,f_0$}},
xlabel style={font=\color{black}},
xlabel={Frequency (Hz)},
ymin=-0.6,
ymax=1.8,
ytick={-1,0,1},
yticklabels={{$-H_{max}$},{$0$},{$H_{max}$}},
ylabel style={font=\color{black}},
ylabel={Velocity Gain (dB)},
axis background/.style={fill=white},
legend style={at={(0.03,0.97)}, anchor=north west, legend cell align=left, align=left, draw=black}
]
\addplot [color=mycolor1, line width=1.0pt]
  table[row sep=crcr]{%
0.144955932735539	0\\
6.89864830730608	0\\
};
\addlegendentry{Rigid}

\addplot [color=mycolor2, line width=1.0pt]
  table[row sep=crcr]{%
0.19863396577163	0.0130853238351267\\
0.258325707549833	0.0224486189235784\\
0.315685558331593	0.0341209791885302\\
0.37010400414833	0.0478766180948162\\
0.422066459018271	0.0637532079254561\\
0.471442671240798	0.0816730341091105\\
0.519362909256897	0.102124343365114\\
0.564295993677053	0.124518414783301\\
0.608891624114775	0.150423485143511\\
0.6524842456087	0.18005859125121\\
0.694379765783376	0.213542966500311\\
0.733873289333382	0.250823019913703\\
0.770268436571818	0.291557560485903\\
0.808468536738449	0.343155544124363\\
0.842715808843405	0.400135714492191\\
0.872360828175746	0.461262412091694\\
0.903048698682816	0.541931248965001\\
0.928374457504588	0.629323978584892\\
0.954410470447576	0.751410188426312\\
1.00174276895567	1\\
1.00869349286323	0.977189375585727\\
1.08840932061445	0.538188698312604\\
1.13451507299787	0.400776610693177\\
1.19077934058484	0.281117297149796\\
1.2585060720474	0.173526395114029\\
1.33931379558443	0.0741844509936445\\
1.44515814631042	-0.0281989563319276\\
1.5701871228339	-0.124770936518976\\
1.72979028403689	-0.224554212349963\\
1.94555944323041	-0.333511415505057\\
2.23411006189384	-0.450534982223141\\
2.65570382810883	-0.586005231283067\\
3.29058211347136	-0.743717046995796\\
4.3390319978701	-0.937490422621834\\
5.01713648781024	-1.03656332339702\\
};
\addlegendentry{Flexible}

\addplot [color=black, forget plot]
  table[row sep=crcr]{%
1	-1.2\\
1	0.99\\
};
\end{axis}
\end{tikzpicture}%
\label{fig:resonance-velocity}}
	\hfill
	\subfloat[Velocity Time Domain Response]{
%
%
\colorlet{mycolor1}{colorRigid}
\colorlet{mycolor2}{colorFlexible}
\begin{tikzpicture}

\begin{axis}[%
width=0.951\figurewidth,
height=\figureheight,
at={(0\figurewidth,0\figureheight)},
scale only axis,
xmin=0,
xmax=1.2,
xtick={0,0.5,1},
xticklabels={{$0$},{$T_0/2$},{$T_0$}},
xlabel style={font=\color{black}},
xlabel={Time (s)},
ymin=-2.5,
ymax=3.5,
ytick={-2,-1,0,1,2,3},
yticklabels={
{$-2\,\vhatin$},
{$-\,\vhatin$},
{$0$},
{$\vhatin$},
{$2\,\vhatin$},
{$3\,\vhatin$}},
ylabel style={font=\color{black}},
ylabel={Velocity},
axis background/.style={fill=white},
legend style={at={(0.03,0.97)}, anchor=north west, legend cell align=left, align=left, draw=black}
]
\addplot [color=mycolor1, line width=1.0pt]
  table[row sep=crcr]{%
-0.000467231789834877	0\\
0.00654124505768794	-0.00187075800137504\\
0.0112135629560366	-0.0149227144964476\\
0.0200909669628988	-0.0613960358474885\\
0.0780277089024211	-0.41295447826522\\
0.11353732492987	-0.603833312228519\\
0.135964450841943	-0.709699326355696\\
0.162129431072695	-0.815568313185542\\
0.185023788774603	-0.889966518425891\\
0.202778596788328	-0.935027583219403\\
0.217730014063043	-0.964103509180459\\
0.232214199547924	-0.984106743334304\\
0.2452966896633	-0.99515418956324\\
0.256510252619336	-0.999300330643955\\
0.267723815575373	-0.998501874733661\\
0.279404610321244	-0.992338845487036\\
0.290618173277281	-0.981531246269093\\
0.305569590551996	-0.959398541603494\\
0.323324398565721	-0.922280186671056\\
0.34341536552862	-0.866247336838345\\
0.366776955020362	-0.783983550472106\\
0.394810862410454	-0.66319178275561\\
0.431722173807407	-0.47364691643243\\
0.492462306485939	-0.111730259049013\\
0.543857803367773	0.209268082403257\\
0.574695101496873	0.393550420451208\\
0.627025061958377	0.66904764934995\\
0.649919419660285	0.768615420541204\\
0.672346545572358	0.850917995998652\\
0.697577062223441	0.922944056249922\\
0.712995711287991	0.955827884597878\\
0.728414360352541	0.979633591065628\\
0.742431314047587	0.99331217074052\\
0.754579340583293	0.998935759657574\\
0.760186122061311	0.999569648541892\\
};
\addlegendentry{Rigid}

\addplot [color=mycolor2, line width=1.0pt]
  table[row sep=crcr]{%
-0.000467231789834877	0\\
0.046255947193651	-0.00199396467543167\\
0.0668141459463842	-0.00749579906815478\\
0.0845689539601091	-0.0167346646091122\\
0.101389298394164	-0.0304313108933973\\
0.118209642828218	-0.0498250806680507\\
0.135497219052108	-0.0764693744267158\\
0.153719258855668	-0.112610097821785\\
0.172875762238897	-0.160074278365329\\
0.193901192781465	-0.223689149785075\\
0.216795550483373	-0.306597703066748\\
0.24249329892429	-0.415695960876846\\
0.272396133473721	-0.561294448567428\\
0.311176372030014	-0.771830739197398\\
0.399950412098637	-1.26083766477322\\
0.426582624119224	-1.38172133569599\\
0.448075286451627	-1.46165878697254\\
0.465830094465352	-1.51299505646123\\
0.480314279950232	-1.543561909059\\
0.492929538275773	-1.56106893706984\\
0.503675869441975	-1.56881078164764\\
0.513020505238672	-1.56989603051747\\
0.521897909245534	-1.56586364715168\\
0.531709776832066	-1.55547049712335\\
0.542456107998268	-1.53671531333622\\
0.554604134533975	-1.50596980110567\\
0.568153856439185	-1.45943509965184\\
0.583572505503736	-1.39048274873953\\
0.600392849937791	-1.29564132420782\\
0.619082121531185	-1.16630516798744\\
0.640107552073753	-0.991259417308592\\
0.663469141565496	-0.76173228979685\\
0.690101353586083	-0.458824468386448\\
0.721405883505018	-0.0547396945535881\\
0.761587817430816	0.519517752426838\\
0.872788983411512	2.13945316562989\\
0.901757354381273	2.48872554666015\\
0.925586175662851	2.73061894082791\\
0.945677142625749	2.89620474858542\\
0.962497487059804	3.00438156311106\\
0.976981672544685	3.07338795438026\\
0.989129699080391	3.11302313995595\\
0.998941566666923	3.13239330369368\\
1.00688450709412	3.1395677734031\\
1.01015512962296	3.14027378459678\\
};
\addlegendentry{Flexible}

\addplot [color=black, forget plot]
  table[row sep=crcr]{%
0.7602	0\\
0.7602	0.99\\
};
\addplot [color=black, forget plot]
  table[row sep=crcr]{%
1.01	0\\
1.01	3.14\\
};
\addplot [color=black, forget plot]
  table[row sep=crcr]{%
0	0\\
1.2	0\\
};
\draw (axis cs:0.25,-1.8) -- (axis cs:0.7602,-1.8);
\draw (axis cs:0.25,-1.7) -- (axis cs:0.25,-1.9);
\draw (axis cs:0.7602,-1.7) -- (axis cs:0.7602,-1.9);
\node[anchor=north]
at (axis cs:0.95,0) {$\Ggain = \frac{\vhat}{\vhatin}$};
\node[anchor=west] at (axis cs:0.761,0.5) {\vhatrigid};
\node[anchor=west] at (axis cs:1,0.5) {\vhatflex};
\node[anchor=west] at (axis cs:0.77,-1.8) {$\frac{1}{2\;\ffrequency}$};
\end{axis}
\end{tikzpicture}%
\label{fig:timedomain-velocity}}
	\hfill
	\subfloat[Energy Time Domain Response]{
%
%
\colorlet{mycolor1}{colorRigid}
\colorlet{mycolor2}{colorFlexible}
\begin{tikzpicture}

\begin{axis}[%
width=0.951\figurewidth,
height=\figureheight,
at={(0\figurewidth,0\figureheight)},
scale only axis,
xmin=0,
xmax=1.2,
xtick={0,0.5,1},
xticklabels={{$0$},{$T_0/2$},{$T_0$}},
xlabel style={font=\color{black}},
xlabel={Time (s)},
ymin=0,
ymax=10.2,
ytick={0,1,2,3,4,5,6,7,8,9,10},
yticklabels={{$0$},
{$\hat{E}_{rig}$},
,
{$3\,\hat{E}_{rig}$},
,
{$5\,\hat{E}_{rig}$},
,
{$7\,\hat{E}_{rig}$},
,
{$9\,\hat{E}_{rig}$},
},
ylabel style={font=\color{black}},
ylabel={Energy},
axis background/.style={fill=white},
legend style={at={(0.03,0.97)}, anchor=north west, legend cell align=left, align=left, draw=black}
]
\addplot [color=mycolor1, line width=1.0pt]
  table[row sep=crcr]{%
-0.000467231789834877	0\\
0.0168203444340549	0.00178429912529987\\
0.0247632848612473	0.00805995262613268\\
0.0345751524477793	0.0228908234701728\\
0.0471904107733205	0.0524525307319716\\
0.0602729008886964	0.0950547330455083\\
0.0728881592142376	0.146264156504944\\
0.0948480533364757	0.255319978946636\\
0.131292132943595	0.472589739223017\\
0.185491020564438	0.791731748620708\\
0.205114755737502	0.881075189355506\\
0.224271259120731	0.945985010456517\\
0.238288212815777	0.977358571939887\\
0.249034543981979	0.991401148957461\\
0.257911947988841	0.996291440436622\\
0.266322120205868	0.995183316377003\\
0.274732292422895	0.988489607995831\\
0.283142464639923	0.97640452752696\\
0.29435602759596	0.952084038870122\\
0.305102358762161	0.920419191475991\\
0.323324398565721	0.849448894576148\\
0.343882597318454	0.747306662484558\\
0.377523286186564	0.548669064446389\\
0.436394491705756	0.202900924229747\\
0.453214836139811	0.125920772060367\\
0.47237133952304	0.0589070451662179\\
0.485921061428251	0.0265740665452128\\
0.497601856174122	0.00975562537760366\\
0.506479260180984	0.00403152469967827\\
0.514889432398012	0.00428001065702954\\
0.523299604615039	0.0101053345060811\\
0.532177008621901	0.0222189139380808\\
0.541988876208433	0.0424404479081741\\
0.552735207374635	0.0725002188263293\\
0.56955555180869	0.135335841168238\\
0.583105273713901	0.19785283947516\\
0.603196240676799	0.305703718724323\\
0.629361220907552	0.464377851008788\\
0.676551631680872	0.748287869548132\\
0.70038045296245	0.86494377292061\\
0.720938651715183	0.939848384608609\\
0.733553910040724	0.971873465221433\\
0.745701936576431	0.991503053744538\\
0.755046572373128	0.998865765529617\\
0.760186122061311	1\\
};
\addlegendentry{Rigid}

\addplot [color=mycolor2, line width=1.0pt]
  table[row sep=crcr]{%
-0.000467231789833988	0\\
0.0579367419395229	0.0019890590318532\\
0.0766260135329162	0.00727825910188251\\
0.0915774308076323	0.0160497495728773\\
0.105594384502677	0.0296256087977067\\
0.119144106407889	0.0491519400958982\\
0.133161060102934	0.0774714138246608\\
0.147645245587814	0.116931699541325\\
0.163063894652366	0.171879020947694\\
0.179417007296585	0.246169608905095\\
0.197171815310309	0.346576775774611\\
0.216795550483374	0.481699697080666\\
0.239222676395446	0.665416370731975\\
0.266789351995703	0.926901523818048\\
0.311643603819849	1.39823256854496\\
0.353694464904986	1.82449899521936\\
0.379392213345904	2.04551907906427\\
0.399950412098637	2.18995205248197\\
0.417705220112362	2.28826723552997\\
0.433591100966746	2.35485654513158\\
0.447608054661792	2.39744302237946\\
0.460690544777169	2.42468785706247\\
0.472838571312876	2.440614757314\\
0.484519366058747	2.44903688055291\\
0.497601856174121	2.45273679889922\\
0.537316558310085	2.45854128512228\\
0.547595657686452	2.46728431137112\\
0.557407525272984	2.48171696958616\\
0.567686624649351	2.50498562376436\\
0.578900187605386	2.54195733294029\\
0.590580982351259	2.59563796952641\\
0.6031962406768	2.67362836933393\\
0.61674596258201	2.78337286494241\\
0.631697379856726	2.93870823894722\\
0.648050492500946	3.15253906634423\\
0.666272532304506	3.44671610521869\\
0.686363499267404	3.83869324725301\\
0.709725088759146	4.37716507024486\\
0.738226227939073	5.13309257555225\\
0.784014943342889	6.47325049311978\\
0.830270890536539	7.78860216309765\\
0.857837566136796	8.46153139770408\\
0.88026469204887	8.91554911277559\\
0.899421195432099	9.22714161729142\\
0.916241539866153	9.44060888561454\\
0.931660188930703	9.58806844429661\\
0.945209910835914	9.68212995641559\\
0.95735793737162	9.7411253227531\\
0.968571500327657	9.77722305902216\\
0.978850599704025	9.79748739915172\\
0.988662467290556	9.807901743579\\
0.998941566666923	9.81226496522261\\
1.01015512962296	9.81301861515933\\
};
\addlegendentry{Flex. Total}

\addplot [color=mycolor2, dashed, line width=1.0pt]
  table[row sep=crcr]{%
-0.000467231789834877	0\\
0.057936741939522	0.0019685743822162\\
0.0766260135329162	0.00713490135198569\\
0.0920446625974662	0.0159208240986057\\
0.106528848082347	0.0294845226566594\\
0.120545801777393	0.0487920818278766\\
0.135029987262273	0.0762548266084266\\
0.150448636326824	0.114814176612326\\
0.167268980760879	0.168393026713441\\
0.186892715933943	0.245646671101055\\
0.21072153721552	0.357376987773657\\
0.287347550748437	0.73163059155133\\
0.302766199812987	0.780907295889522\\
0.314914226348693	0.807294357103336\\
0.324726093935225	0.819442038778192\\
0.332669034362418	0.822794974549272\\
0.340144742999775	0.820408536269992\\
0.348087683426968	0.811830582484292\\
0.356965087433831	0.794790645374059\\
0.367711418600032	0.763727125451668\\
0.380326676925573	0.713259246412941\\
0.395278094200289	0.635707785325756\\
0.414901829373353	0.510714502949166\\
0.479379816370563	0.0801847060927638\\
0.491995074696104	0.0298028210575536\\
0.501339710492801	0.00739824432899994\\
0.508348187340323	0.000345351175538688\\
0.513954968818342	0.0013530263212207\\
0.520028982086195	0.00962443776115229\\
0.526570227143883	0.0274242213314579\\
0.534980399360911	0.0647283470800932\\
0.544325035157607	0.12636731095685\\
0.555538598113644	0.229958360296527\\
0.568153856439185	0.386753270742547\\
0.582638041924066	0.620428219582216\\
0.598991154568286	0.952226932697475\\
0.618147657951515	1.42632152841525\\
0.641976479233093	2.12315861886496\\
0.680289485999551	3.38365585599751\\
0.717200797396504	4.55374002004394\\
0.738226227939073	5.10099196122586\\
0.754579340583293	5.42932125718537\\
0.767194598908834	5.61139565648689\\
0.777473698285201	5.70848017221637\\
0.785416638712394	5.74990096494607\\
0.791023420190412	5.76086836744692\\
0.795228506298925	5.75897654258092\\
0.799900824197274	5.74659550182655\\
0.805974837465127	5.71421654568327\\
0.813450546102485	5.6490383771558\\
0.822795181899182	5.52855305599403\\
0.834008744855218	5.32811107277598\\
0.84755846676043	5.0088931573475\\
0.863911579404649	4.52294330864225\\
0.884937009947218	3.76880288863084\\
0.921848321344171	2.26787543677201\\
0.951751155893603	1.12350543404506\\
0.969973195697162	0.566863741261623\\
0.983990149392207	0.251206734983483\\
0.99473648055841	0.0902060084099929\\
1.0026794209856	0.0219177887738375\\
1.00828620246362	0.00150966550950571\\
1.01015512962296	5.03883841496133e-06\\
};
\addlegendentry{Flex. Pot.}

\addplot [color=mycolor2, dashdotted, line width=1.0pt]
  table[row sep=crcr]{%
-0.000467231789833988	0\\
0.114004556719705	0.00196168106118222\\
0.139235073370788	0.0068865397124096\\
0.157924344964181	0.0148554849533138\\
0.173810225818567	0.0263236123486763\\
0.188761643093281	0.0426475696694411\\
0.203245828578163	0.0651313963270468\\
0.217730014063044	0.0957996178841078\\
0.232681431337758	0.137808937910746\\
0.248100080402308	0.194037750422581\\
0.264453193046529	0.270003414206011\\
0.281740769270419	0.370521299813284\\
0.300430040863812	0.504088113450425\\
0.320988239616547	0.681256776479918\\
0.344817060898125	0.92352892849946\\
0.375187127237389	1.27662481541974\\
0.44153404139394	2.06129332114382\\
0.461625008356839	2.24542154091624\\
0.477043657421389	2.35265445037432\\
0.489191683957095	2.41177367262471\\
0.499003551543627	2.44121784698348\\
0.506012028391149	2.45152573508078\\
0.511618809869168	2.45306659173802\\
0.517225591347186	2.44848911057403\\
0.523766836404874	2.43526649990313\\
0.531709776832066	2.40763655590461\\
0.541054412628764	2.35877652038982\\
0.552267975584801	2.27699451328343\\
0.565817697490012	2.14560914704446\\
0.581703578344396	1.94971629248267\\
0.601794545307294	1.64777596865443\\
0.631697379856726	1.12859277777165\\
0.670010386623185	0.475442653647749\\
0.688232426426744	0.230613511713662\\
0.701782148331954	0.0978326716107549\\
0.712061247708322	0.0322911429856489\\
0.719536956345678	0.00638688266825582\\
0.724676506033862	9.94266296814317e-05\\
0.728881592142375	0.00230616697628783\\
0.733553910040724	0.0128187036131013\\
0.739627923308579	0.0396363285776449\\
0.747103631945935	0.0938438905233472\\
0.755981035952798	0.189790734858111\\
0.766727367119	0.353405658983943\\
0.77934262544454	0.613418637994654\\
0.793826810929421	1.00284619344958\\
0.81017992357364	1.55605321039436\\
0.829803658746705	2.36600584082691\\
0.854099711818117	3.55033474710626\\
0.892879950374411	5.67228372398177\\
0.931192957140869	7.6937963200172\\
0.952685619473272	8.62942854892919\\
0.969505963907327	9.20029774629317\\
0.983055685812538	9.5341690973571\\
0.993802016978739	9.70957515880989\\
1.00174495740593	9.78509586474358\\
1.00735173888395	9.80973611387558\\
1.01015512962296	9.81301357632092\\
};
\addlegendentry{Flex. Kin.}

\addplot [color=black, forget plot]
  table[row sep=crcr]{%
0.7602	0\\
0.7602	1\\
};
\addplot [color=black, forget plot]
  table[row sep=crcr]{%
1.01	0\\
1.01	9.81\\
};
\node[anchor=west] at (axis cs:0.761,0.5) {\Ehatrigid};
\node[anchor=west] at (axis cs:1,0.5) {\Ehatflex};
\end{axis}
\end{tikzpicture}%
\label{fig:timedomain-energy}}
	}
	\caption{Comparison of rigid hammer and flexible hammer driven by an ideal velocity source:
	\protect\subref{fig:rigid-tool-model} rigid hammer model,
	\protect\subref{fig:flexible-tool-model} flexible hammer model,
	\protect\subref{fig:resonance-velocity} velocity transfer function magnitude plot,
	\protect\subref{fig:timedomain-velocity} velocity response to a hammering motion in the time domain until impact (vertical black line),
	\protect\subref{fig:timedomain-energy} energy response to a hammering motion in the time domain until impact (vertical black line).
	\mmass is the mass of the hammerhead;
	\bdamping is the damping factor;
	\kstiffness is the spring stiffness;
	\vin is the source velocity;
	\vrig and \vflx are the velocities of the rigid and the flexible hammer, respectively;
	\fresonance is the resonance frequency;
	\ensuremath{\Hmax = \max{\abs{\Hflexible(\slap)}}} is the maximum gain of the flexible hammer;
	\Tresonance = 1/\fresonance	is the resonance period;
	\vhatrigid and \vhatflex are the peak velocities of the rigid and the flexible hammer;
	$\Ggain = \frac{\vhat}{\vhatin}$ is the velocity gain, with $\vhat = \vhatrigid$ for the rigid hammer and $\vhat = \vhatflex$ for the flexible hammer;
	\ffrequency is the excitation frequency;
	\Ehatrigid and \Ehatflex are the peak energy of the rigid and the flexible hammer, respectively.%
	}
	\label{fig:resonance-analysis}
\end{figure*}

The frequency-domain responses and time domain responses of rigid and flexible hammers can be analyzed in detail using equations \ref{eq:Hrigid}, \ref{eq:Hflexible}, and \ref{eq:fresonance}:
Fig.~\ref{fig:resonance-velocity} shows the velocity transfer function magnitudes of the rigid hammer $\norm{\Hrigid(\slap)}$ and the flexible hammer $\norm{\Hflexible(\slap)}$ over the frequency \ffrequency.
$\norm{\Hrigid(\slap)}$ equals 1 for all frequencies, meaning that the amplitude of the output velocity $\vrig(\ttime)$ of the rigid tool always equals the amplitude of the input velocity $\vin(\ttime)$.
However, $\norm{\Hflexible(\slap)}$ is considerably greater than 1 for frequencies around the resonance frequency, meaning that the amplitude of the output velocity $\vflx(\ttime)$ is much higher than that of the input velocity $\vin(\ttime)$ at these frequencies.
Only for excitation frequencies considerably higher than the resonance frequency (for frequencies greater than approximately $1.5\,\fresonance$), is the output velocity of the flexible tool lower than the input velocity.

Fig.~\ref{fig:resonance-velocity} shows that the excitation frequency should be as close as possible to the resonance frequency \fresonance in order to achieve the highest velocity with a flexible hammer.
This holds for a periodic motion, as well as for a motion with only one reversal of motion, like a single hammer strike, as will be analyzed next.

\subsection{Implications of Mechanical Resonance for Hammering with a Flexible Hammer}
The principle of hammering is to impact a hammerhead on a target with maximal kinetic energy \Ekinetic.
\Ekinetic depends on the mass of the hammerhead \mmass and its velocity \vvelocity (equal to \vrig for the rigid hammer or \vflx for the flexible hammer), as
\begin{dmath}
\Ekinetic = \frac{1}{2}\,\mmass\,\vvelocity^2
\end{dmath}.
Fig.~\ref{fig:timedomain-velocity} shows the velocity response in the time domain to a hammering motion at the resonance frequency of the flexible hammer compared to the response of the rigid hammer driven by the same velocity source (the input velocity is equal to the velocity of the rigid tool and therefore not displayed in the plot).

In the following, we will refer to the motion profile depicted in Fig.~\ref{fig:timedomain-velocity} as one \emph{strike}.
The strike starts with a backward motion (negative velocity) followed by a forward motion (positive velocity) until the peak, marked by the vertical line.
The peak should coincide with the impact when hammering on a target.
The reversal of motion direction has to occur after half the resonance period \Tresonance for the excitation frequency \ffrequency to match the resonance frequency \fresonance.
We refer to this as the optimal timing: 
\begin{dmath}[compact]
\Topt = \frac{\Tresonance}{2} = \frac{1}{2\,\fresonance} = \pi\sqrt{\frac{m}{k}}
\end{dmath}.
Hence, the ability to accurately time the reversal is critical to achieve maximum impact with a flexible hammer.

Fig.~\ref{fig:timedomain-energy} shows the energy increase in the flexible hammer compared to the rigid hammer during one strike.
While the energy of the rigid hammer consists only of the kinetic energy \Ekinetic, the energy of the flexible hammer equals to the sum of the kinetic energy and the potential energy in the spring \Espring:
\begin{dmath}
\Espring = \frac{1}{2}\,\kstiffness\,\xdelta^2
\end{dmath},
with \xdelta being the spring deflection.
If the excitation frequency equals the resonance frequency, the kinetic energy of the hammerhead is fully transferred into potential energy of the spring during the reversal of motion and back into kinetic energy during the forward swing motion, thereby achieving the highest kinetic energy on impact.
Thus, exciting the mechanical resonance of a flexible hammer increases the power transferred to the hammerhead, which results in a significantly increased peak velocity and peak energy compared to a rigid hammer driven by the same velocity source.

\section{\ExpOne: Direct Manipulation}
\label{sec:experiment-1}
\ExpOne was previously presented in more detail in \cite{Aiple2017}. The results were reprocessed in this study to allow for a comparison with the results of \ExpTwo (described later in Section~\ref{sec:experiment-2}).
\subsection{Hypothesis}
The following hypothesis was formulated for the direct manipulation experiment: 

\medskip
\noindent \emph{Hypothesis H1}: Humans can exploit the elasticity of a flexible hammer in direct manipulation to maximize power transfer in a strike, and to achieve higher output velocity than with a rigid hammer.

\subsection{Conditions}
Five hammer stiffness conditions were used (with corresponding resonance frequencies): \SI{0.62}{\N\m\per\radian} (\SI{3}{\Hz}), \SI{2.3}{\N\m\per\radian} (\SI{4.8}{\Hz}), \SI{4.1}{\N\m\per\radian} (\SI{6.9}{\Hz}), \SI{11}{\N\m\per\radian} (\SI{9.9}{\Hz}), and rigid (stiffness higher than \SI{10000}{\N\m\per\radian} and resonance frequency higher than \SI{300}{\Hz}). The conditions are denoted in the text using the resonance frequency (determined through system identification measurements \cite{Aiple2017}).
The stiffness settings were chosen based on a pilot study, where for stiffness settings below \SI{0.62}{\N\m\per\radian}, pilot participants reported that they did not get the impression that they were creating any impact. For stiffness settings higher than \SI{11}{\N\m\per\radian}, pilot participants reported not being able to distinguish between flexible and rigid hammer extension.

\subsection{Participants}
The experiment was performed by 13 participants (8 male, 5 female, 12 righthanded, 1 ambidextrous, age 21-41).
The participants were researchers and project engineers with a higher education background.
None of the participants had prior experience with the experiment.
The experiment had been approved by the human research ethics committee of TU Delft and all participants gave written informed consent before participation.

\subsection{Experiment Procedure}
\begin{figure}[!t]
	\footnotesize
\begin{tikzpicture}
[box/.style={draw},
fambox/.style={box},
condbox/.style={box,anchor=west,minimum height=0.7cm,minimum width=1.5cm},
phasebox/.style={box,anchor=west,minimum height=1cm,align=center}
]
\node[condbox] (cond1) at (0,0cm){Condition};
\foreach \i in {2,...,5}
{
\node[condbox] (cond\i) at ({(\i-1)*1.5cm},0cm){\dots};
}
\node[phasebox,minimum width=4cm] (phasetrain) at ($(cond1.south west)+(0.5cm,-1cm)$) {Training \\ 100 trials};
\node[phasebox,minimum width=2cm] (phaseperf) at ($(cond1.south west)+(5.5cm,-1cm)$) {Performance \\ 20 trials};
\node[phasebox,minimum width=1cm] (phasebreak) at ($(cond1.south west)+(4.5cm,-1cm)$) {Break};
\draw[dashed] (cond1.south west) -- (phasetrain.north west);
\draw[dashed] (cond1.south east) -- (phaseperf.north east);
\draw[->,>=latex] (phaseperf.south) -- ($(phaseperf.south)+(0cm,-0.8cm)$);
\node[fill=white] (regr) at ($(phaseperf.south)+(0cm,-0.35cm)$) {10 Highest Gain Trials};
\begin{scope}[shift={($(phaseperf.south)+(0cm,-0.8cm)$)}]
\node[fill=white] (b) at (0cm,-1em) {$\bvalue{x} = \text{median}(x)$};
\end{scope}
\draw[->,>=latex] ($(cond1.north west)+(0cm,0.3cm)$) -- (cond1.north west);
\node [anchor=south] (instr) at ($(cond1.north west)+(0cm,0.3cm)$) {Instructions};
\foreach \i in {2,...,5}
{
\draw[->,>=latex] ($(cond\i.north west)+(0cm,0.3cm)$) -- (cond\i.north west);
\node [anchor=south] (break\i) at ($(cond\i.north west)+(0cm,0.3cm)$) {Break};
}
\end{tikzpicture}
	\caption{Experimental procedure of \ExpOne.}
	\label{fig:conditions-exp1}
\end{figure}
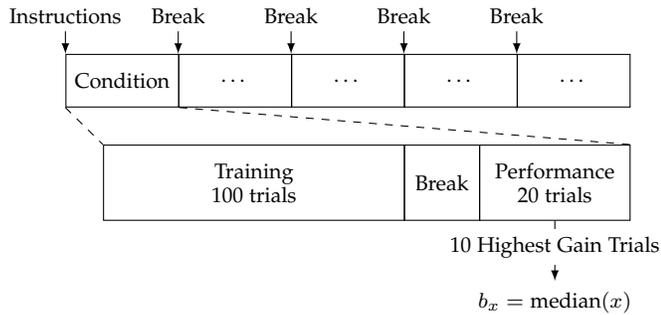

Fig.~\ref{fig:conditions-exp1} visualizes the experiment procedure. The participants received verbal task instructions and then conducted 100~training trials followed by 20~performance trials per condition.
The order of conditions was randomly selected for each participant.
Between training trials and performance trials, there was a short break (approximately \SI{1}{\minute}) to download the data and between two conditions there was a break to download the data and change the spring of the flexible hammer extension (approximately~\SI{5}{\minute}).
Only the trials conducted in the performance phase were used for the analysis.

\subsection{Experiment Apparatus}
\label{sec:experiment-setup}
\begin{figure}[!t]
	\centering
	\subfloat[Direct manipulation]{\includegraphics[height=4cm]{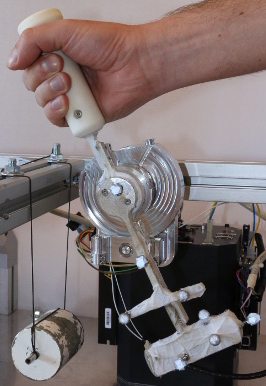}}
    \hspace{2mm}
	\subfloat[Teleoperated manipulation]{\includegraphics[height=4cm]{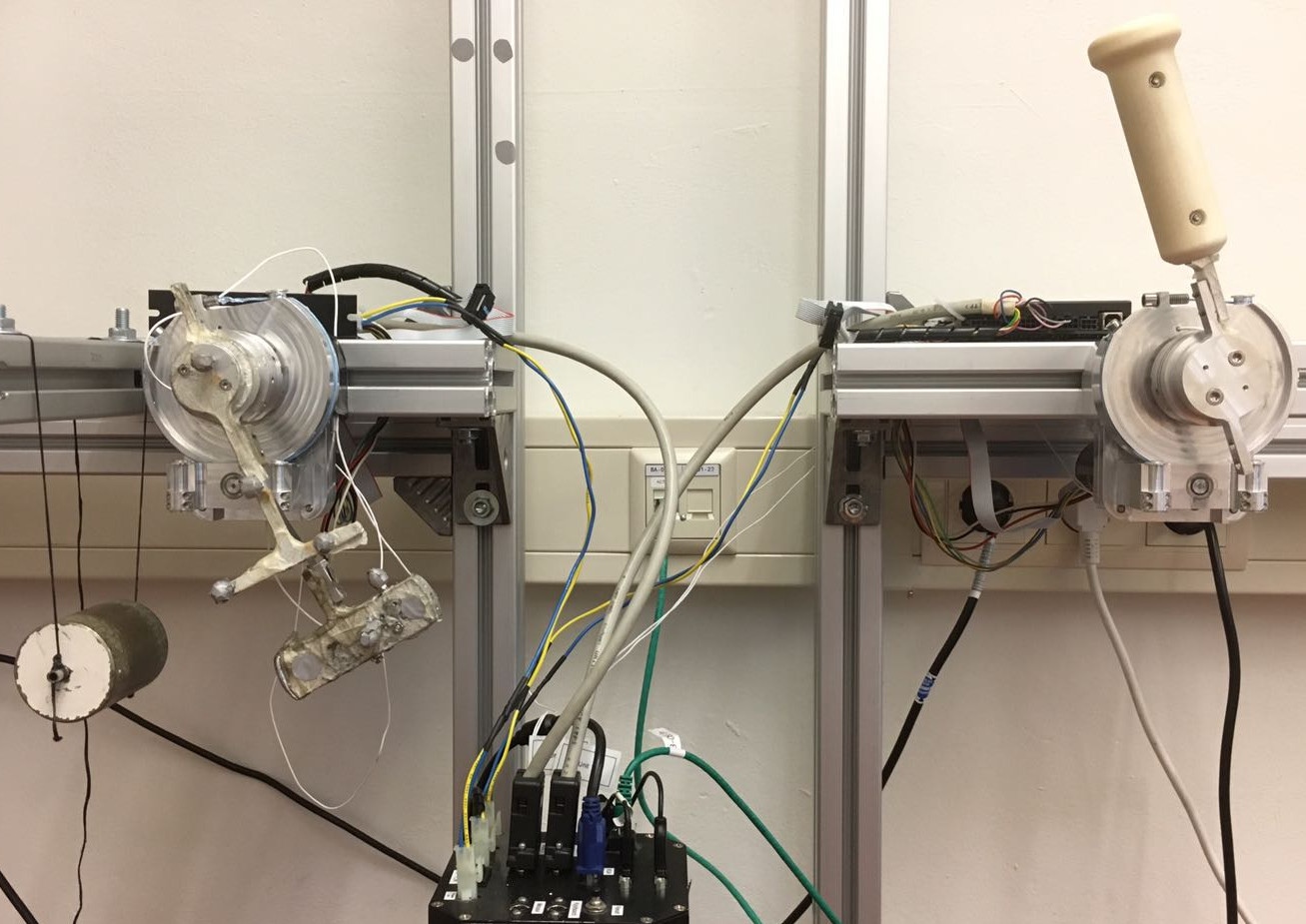}}
	\caption{Experimental apparatus used for the two experiments.}
	\label{fig:experiment-setups}
\end{figure}

\begin{figure*}[!t]
	\centering
	\footnotesize \input{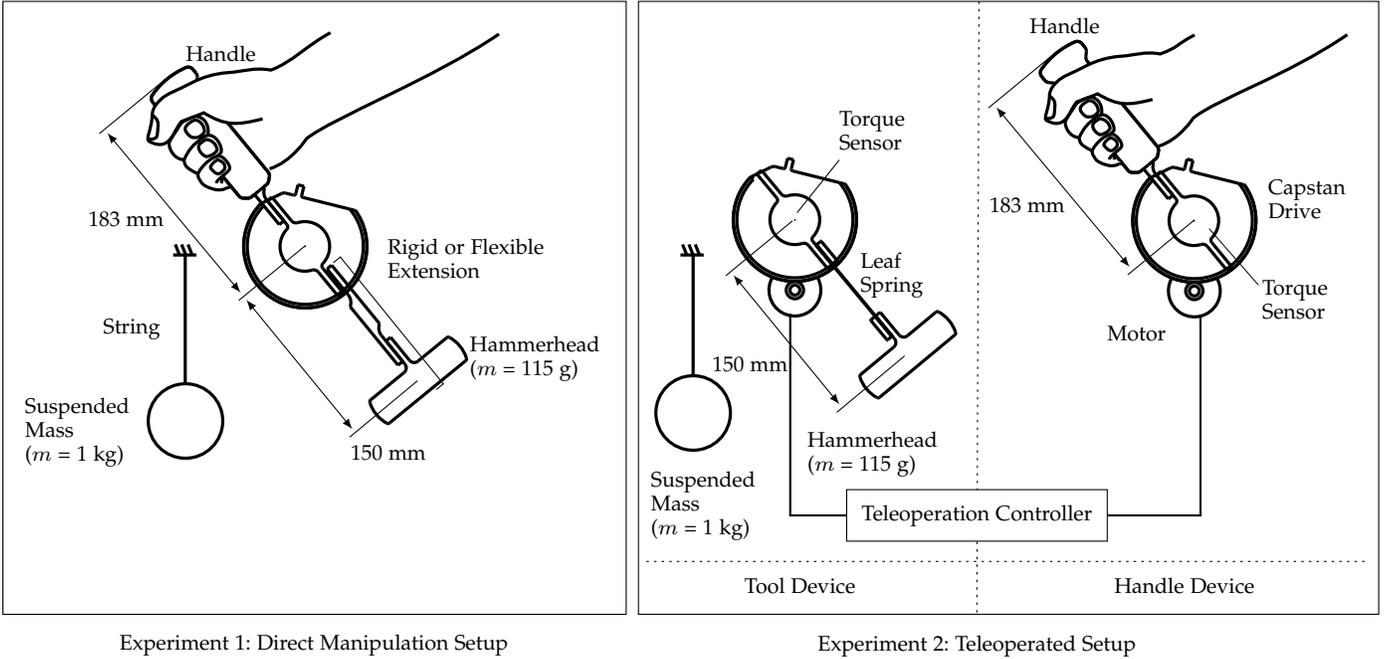}
	\caption{Schematic view  of the setups for \ExpOne and \ExpTwo.
			In \ExpOne, the handle is mounted on the same rotating axis as the hammer.
			In \ExpTwo, the handle is mounted to the handle device, whereas the hammer is mounted to the tool device.
			Handle device and tool device are coupled by a four-channel teleoperator.
			They are fixed to separate frames to isolate the vibrations between tool device and handle device.}
	\label{fig:combined-elastic-setup}
\end{figure*}

An apparatus was designed for this study to enable the performance of a one degree of freedom flexible hammering task in either direct or teleoperated manipulation with identical handle and tool devices (see photos in Fig.~\ref{fig:experiment-setups} as well as the illustration in Fig.~\ref{fig:combined-elastic-setup}).
Although the required motion of the operator is different from hammering with a conventional hammer, the authors believe that the setup provides good insights into how humans interact with flexible tools in general.
The teleoperation system used in \ExpTwo is explained in more detail in Section~\ref{sec:experiment-2}.

The handle device and the tool device had a one degree of freedom rotational actuator with torque sensor, driven by a \SI{70}{\W} brushless DC Maxon EC-max 40 motor with gear and capstan drive (total gear ratio 35:1).
It was a modified version of the setup described by Rebelo et al. \cite{J.Rebelo2015}.
The hammer consisted of a \SI{115}{\g} hammerhead (similar to light commercial hammers) attached to the output shaft of the tool device through an exchangeable extension in such a way that the distance between rotation axis and hammerhead center of mass was \SI{150}{\mm}.
The extension consisted of either a leaf-spring or a rigid extension of the same length.
For simplification, linear behavior of the leaf-spring was assumed in general.
However, the methods used for data processing and analysis did not require linearity of the spring.
The handle device and the tool device were mounted on separate mechanical structures to prevent any vibrations from being propagated from the tool device to the handle device.

A handle for the operator was mounted on the output shaft of the tool device or handle device with a distance of \SI{183}{\mm} between rotation axis and handle tip.
In the direct manipulation configuration, the handle was mounted to the tool device and there was a mechanical coupling between the handle held by the operator and the hammer.
In this configuration, the motor of the tool device was decoupled from the capstan drive to reduce friction.

In the teleoperated manipulation configuration, the handle was mounted on the handle device and the tool device was driven by the motor controlled by the four-channel bilateral teleoperation controller running at \SI{1}{\kHz} (see section~\ref{sec:teleoperator-tuning}).
At the same time, force feedback was provided to the operator through the motor on the handle device side, which was also controlled by the teleoperation controller.

A mass of \SI{1}{\kg} served as the target to be struck by the flexible hammer.
It was suspended in such a way that it could be struck by the hammer and provide the participants with a simple visual impression of the impact energy:
The greater the impact energy, the higher the target swung after the impact. 

\subsection{Experiment Task}
\label{sec:experiment-task}
The task for both experiments was identical: a one degree of freedom flexible hammering task.
The motion should consist of one hammer strike per trial consisting of one backward swing followed by one forward swing.
The goal of the task for the participants was to find the best timing for achieving maximum impact with least effort, while using a similar effort in every trial and focusing primarily on the timing of the change of direction of motion.
It was communicated to the participants that what was important was the velocity gain, not the absolute energy in the impact.
In other words, the task instructions were explicitly formulated with the emphasis on the velocity gain and timing for resonance, because during pilot studies participants had tended to put all their effort into achieving maximum impact, thereby quickly suffering fatigue without utilizing the resonance (which was not in accordance with the goal of the experiment).

\subsection{Data Acquisition}
\label{sec:acquisition-analysis}
The positions of the tool device joint and hammerhead were tracked simultaneously using one Vicon motion capture system at a sampling rate of \SI{1}{\kHz}. To this end, reflective markers were attached rigidly to the actuator joint and to the hammerhead.
This way, all relevant data was acquired by the same system and inaccuracies through the measurement method were reduced.
In the post-processing, these positions were transformed into polar coordinates, the angular velocities of joint and hammer were derived, and the individual trials were segmented and synchronized in such a way that the moment of impact was at time
\begin{math}\ttime = \SI{0}{\s}\end{math}~\cite{Aiple2017}.

\subsection{Dependent Measures}
\label{sec:dependent-variables}
Three dependent measures, listed in Table~\ref{tbl:dependent-measures}, were used to compare the hammering task performance.
As illustrated in Fig.~\ref{fig:timedomain-velocity}, the measures were extracted from the time domain data as following for each segment (cf. \cite{Aiple2017}).
The peak joint velocity \vhatin and peak hammerhead velocity \vhat were obtained by calculating the maximum of the joint velocity \vin and of the hammer velocity \vvelocity before the impact.
The gain $\Ggain = \vhat/\vhatin$ was calculated as the ratio of peak joint velocity to peak hammerhead velocity.
The excitation frequency $\ffrequency = 1/(2 \cdot \Tperiod/2)$ was calculated from half the period $\Tperiod/2$, which was calculated from the time between the minimum and the maximum of the joint velocity \vin.

It was decided to use the peak velocity as measure instead of the impact velocity, which might be different, as the primary research question of this study was how to exploit mechanical resonance and therefore the peak velocity is more relevant than the impact velocity.
In practice, the peak velocity of a given velocity profile can be made to occur at the impact position by adapting the start position.

The peak hammer velocity \vhat measured the achieved result and gave an indication of the effectiveness of the task execution.
The dimensionless gain \Ggain gave an indication of the energy efficiency of the task execution and was the primary measure to be maximized.
The excitation frequency \ffrequency was used to assess whether the participants tried to sense and excite the mechanical resonance of the flexible hammer to achieve better performance.
The closer the excitation frequency to the resonance frequency, the better they had adapted to the system.
Therefore, the excitation frequency indicated whether the participants tried to influence the outcome of the experiment by changing their strategy (hammering faster or slower), or whether they kept the same strategy throughout the experiment and the outcome was influenced only by the properties of the system.

\begin{table}[!t]
	\begin{threeparttable}
		\caption{Dependent Measures}
		\label{tbl:dependent-measures}
		\renewcommand*{\arraystretch}{1.8}
		\begin{tabular}{lcl}
			\toprule
			 Measure & Symbol & Description \\
			\midrule
			Output & \vhat & Peak hammer velocity (\SI{}{\radian\per\s}) \\
			Efficiency & \Ggain & Gain (-) \\
			Adaptation & \ffrequency & Excitation frequency (\SI{}{\Hz}) \\
			\bottomrule
		\end{tabular}
	\end{threeparttable}
\end{table}

\subsection{Data Analysis}
Through the described procedure, a single value per trial of each dependent measure was obtained;
trials were discarded if values for the dependent measures could not be extracted.
For the remaining trials, the 10 best trials were selected based on the highest gain \Ggain and from these 10 trials the median of each measure \vhat, \Ggain, \ffrequency was calculated and notated \bvalue{\vhat}, \bvalue{\Ggain}, and \bvalue{\ffrequency}, respectively.
This was done to ensure that learning or fatigue effects did not affect the results.
In a preceding analysis, the data were analyzed for patterns in the learning or fatigue, correlating with the order or the type of conditions, but none were observed.

\subsection{Statistical Evaluation}
\label{sec:statistical-evaluation}
Friedman's test \cite{Friedman1937TheVariance} was used to check for column effects in the dependent measures (\bvalue{\vhat}, \bvalue{\Ggain} and \bvalue{\ffrequency}) after adjusting for possible row effects ($\alphavar = \SI{5}{\percent}$, row effects: participant-dependent variations, column effects: condition-dependent variations).
For \ExpOne, where Friedman's test indicated an effect, Wilcoxon's signed-rank test \cite{Wilcoxon1945IndividualMethods} was applied for pairwise testing of the conditions ($\alphavar = \SI{1}{\percent}$ to compensate for multiple comparisons). The effect sizes, reported as $\Delta\bvalue{\vhat}$, $\Delta\bvalue{\Ggain}$, and $\Delta\bvalue{\ffrequency}$, were calculated as median of the pair differences, with corresponding 95\%~confidence intervals.
The \SI{4.8}{\Hz} condition of \ExpOne was included as reference condition in the box plots of \ExpTwo for visual comparison of the performance differences between teleoperated manipulation (\ExpTwo) and direct manipulation (\ExpOne).
It was not used for the statistical analysis of \ExpTwo.

Finally, an estimate of the minimum effect size \cite{Murphy2009StatisticalTests} was provided for each measure.
It was calculated through statistical power analysis as the minimum required difference between medians to reject the null hypothesis with $1-\beta=$~\SI{80}{\percent}, given the number of participants and the estimated variation~\cite{Wilcox2001FundamentalsAccuracy}.
Because the data were not normally distributed, the calculation was based on an equivalent standard deviation estimation \sigmastdest using the median absolute deviation \madfunc:
\begin{math}
\sigmastdest = 1.48\,\madfunc
\end{math}~\cite{Ellison2009PracticalGuide.}. 
Thus, the estimates give an indication of the required effect sizes, but an exact comparison with the results of Friedman's test and Wilcoxon's signed-rank test would be difficult to make. 

\subsection{Results}
\begin{figure*}[!t]
	{
	\footnotesize
	\setlength{\figurewidth}{.27\textwidth}
	\setlength{\figureheight}{\plotheightA}
	\subfloat[Peak Hammer Velocity]%
	{
%
%
\colorlet{mycolor1}{color3p0Hz}
\colorlet{mycolor2}{color4p8Hz}
\colorlet{mycolor3}{color6p9Hz}
\colorlet{mycolor4}{color9p9Hz}
\colorlet{mycolor5}{colorRig}
\begin{tikzpicture}

\begin{axis}[%
width=0.951\figurewidth,
height=\figureheight,
at={(0\figurewidth,0\figureheight)},
scale only axis,
xmin=0.5,
xmax=5.5,
xtick={1,2,3,4,5},
xticklabels={{3.0 Hz},{4.8 Hz},{6.9 Hz},{9.9 Hz},{Rigid}},
xlabel style={font=\color{black}},
xlabel={Condition (-)},
ymin=0,
ymax=50,
ylabel style={font=\color{black}},
ylabel={Peak Hammer Velocity (rad/s)},
axis background/.style={fill=white},
axis x line*=bottom,
axis y line*=left
]

\addplot[area legend, line width=2.0pt, draw=black, fill=mycolor1, forget plot]
table[row sep=crcr] {%
x	y\\
0.6	19.0253649840241\\
1.4	19.0253649840241\\
1.4	17.7839476626446\\
1.2	22.013152918867\\
1.4	24.3997226217239\\
1.4	23.4207445392606\\
0.6	23.4207445392606\\
0.6	24.3997226217239\\
0.8	22.013152918867\\
0.6	17.7839476626446\\
0.6	19.0253649840241\\
}--cycle;
\addplot [color=black, line width=2.0pt, forget plot]
  table[row sep=crcr]{%
0.800000000000001	22.013152918867\\
1.2	22.013152918867\\
};
\addplot [color=black, line width=2.0pt, forget plot]
  table[row sep=crcr]{%
1	11.4561270051257\\
1	19.0253649840241\\
};
\addplot [color=black, line width=2.0pt, forget plot]
  table[row sep=crcr]{%
1	23.4207445392606\\
1	34.8799940429922\\
};
\addplot [color=black, line width=2.0pt, forget plot]
  table[row sep=crcr]{%
0.9	11.4561270051257\\
1.1	11.4561270051257\\
};
\addplot [color=black, line width=2.0pt, forget plot]
  table[row sep=crcr]{%
0.899999999999999	34.8799940429922\\
1.1	34.8799940429922\\
};

\addplot[area legend, line width=2.0pt, draw=black, fill=mycolor2, forget plot]
table[row sep=crcr] {%
x	y\\
1.6	26.693859299537\\
2.4	26.693859299537\\
2.4	26.5768412015772\\
2.2	29.8853051935359\\
2.4	36.1106103836636\\
2.4	35.3884466096858\\
1.6	35.3884466096858\\
1.6	36.1106103836636\\
1.8	29.8853051935359\\
1.6	26.5768412015772\\
1.6	26.693859299537\\
}--cycle;
\addplot [color=black, line width=2.0pt, forget plot]
  table[row sep=crcr]{%
1.8	29.8853051935359\\
2.2	29.8853051935359\\
};
\addplot [color=black, line width=2.0pt, forget plot]
  table[row sep=crcr]{%
2	17.3049777442583\\
2	26.693859299537\\
};
\addplot [color=black, line width=2.0pt, forget plot]
  table[row sep=crcr]{%
2	35.3884466096858\\
2	45.8902789022242\\
};
\addplot [color=black, line width=2.0pt, forget plot]
  table[row sep=crcr]{%
1.9	17.3049777442583\\
2.1	17.3049777442583\\
};
\addplot [color=black, line width=2.0pt, forget plot]
  table[row sep=crcr]{%
1.9	45.8902789022242\\
2.1	45.8902789022242\\
};

\addplot[area legend, line width=2.0pt, draw=black, fill=mycolor3, forget plot]
table[row sep=crcr] {%
x	y\\
2.6	24.0435841628804\\
3.4	24.0435841628804\\
3.4	22.7237056150842\\
3.2	30.4866896490035\\
3.4	37.4041185835121\\
3.4	37.3260638817048\\
2.6	37.3260638817048\\
2.6	37.4041185835121\\
2.8	30.4866896490035\\
2.6	22.7237056150842\\
2.6	24.0435841628804\\
}--cycle;
\addplot [color=black, line width=2.0pt, forget plot]
  table[row sep=crcr]{%
2.8	30.4866896490035\\
3.2	30.4866896490035\\
};
\addplot [color=black, line width=2.0pt, forget plot]
  table[row sep=crcr]{%
3	17.8188264050934\\
3	24.0435841628804\\
};
\addplot [color=black, line width=2.0pt, forget plot]
  table[row sep=crcr]{%
3	37.3260638817048\\
3	46.4416687741171\\
};
\addplot [color=black, line width=2.0pt, forget plot]
  table[row sep=crcr]{%
2.9	17.8188264050934\\
3.1	17.8188264050934\\
};
\addplot [color=black, line width=2.0pt, forget plot]
  table[row sep=crcr]{%
2.9	46.4416687741171\\
3.1	46.4416687741171\\
};

\addplot[area legend, line width=2.0pt, draw=black, fill=mycolor4, forget plot]
table[row sep=crcr] {%
x	y\\
3.6	20.7215868449583\\
4.4	20.7215868449583\\
4.4	20.443255413436\\
4.2	25.2649310563288\\
4.4	30.5054317246666\\
4.4	29.4284837371981\\
3.6	29.4284837371981\\
3.6	30.5054317246666\\
3.8	25.2649310563288\\
3.6	20.443255413436\\
3.6	20.7215868449583\\
}--cycle;
\addplot [color=black, line width=2.0pt, forget plot]
  table[row sep=crcr]{%
3.8	25.2649310563288\\
4.2	25.2649310563288\\
};
\addplot [color=black, line width=2.0pt, forget plot]
  table[row sep=crcr]{%
4	19.0994566733216\\
4	20.7215868449583\\
};
\addplot [color=black, line width=2.0pt, forget plot]
  table[row sep=crcr]{%
4	29.4284837371981\\
4	38.1541947579632\\
};
\addplot [color=black, line width=2.0pt, forget plot]
  table[row sep=crcr]{%
3.9	19.0994566733216\\
4.1	19.0994566733216\\
};
\addplot [color=black, line width=2.0pt, forget plot]
  table[row sep=crcr]{%
3.9	38.1541947579632\\
4.1	38.1541947579632\\
};

\addplot[area legend, line width=2.0pt, draw=black, fill=mycolor5, forget plot]
table[row sep=crcr] {%
x	y\\
4.6	16.1877228203706\\
5.4	16.1877228203706\\
5.4	15.9598641670597\\
5.2	18.2806995418185\\
5.4	19.6529539633013\\
5.4	19.5772691084733\\
4.6	19.5772691084733\\
4.6	19.6529539633013\\
4.8	18.2806995418185\\
4.6	15.9598641670597\\
4.6	16.1877228203706\\
}--cycle;
\addplot [color=black, line width=2.0pt, forget plot]
  table[row sep=crcr]{%
4.8	18.2806995418185\\
5.2	18.2806995418185\\
};
\addplot [color=black, line width=2.0pt, forget plot]
  table[row sep=crcr]{%
5	12.2323361764423\\
5	16.1877228203706\\
};
\addplot [color=black, line width=2.0pt, forget plot]
  table[row sep=crcr]{%
5	19.5772691084733\\
5	20.807750938612\\
};
\addplot [color=black, line width=2.0pt, forget plot]
  table[row sep=crcr]{%
4.9	12.2323361764423\\
5.1	12.2323361764423\\
};
\addplot [color=black, line width=2.0pt, forget plot]
  table[row sep=crcr]{%
4.9	20.807750938612\\
5.1	20.807750938612\\
};
\addplot [color=black, forget plot]
  table[row sep=crcr]{%
0.5	0\\
5.5	0\\
};
\draw[line width=0.5mm] (axis cs:2,3) -- (axis cs:2,2) -- (axis cs:5,2) -- (axis cs:5,3);
\draw[line width=0.5mm] (axis cs:3,5) -- (axis cs:3,4) -- (axis cs:5,4) -- (axis cs:5,5);
\draw[line width=0.5mm] (axis cs:4,7) -- (axis cs:4,6) -- (axis cs:5,6) -- (axis cs:5,7);
\draw[line width=0.5mm] (axis cs:1,9) -- (axis cs:1,8) -- (axis cs:2,8) -- (axis cs:2,9);
\node (ref) at (axis cs:2,48) {\scalebox{1}{Ref.}};
\end{axis}
\end{tikzpicture}%
\label{fig:stats-hammer-velocity-all-exp1-friedman}}
	\hfill
	\subfloat[Gain]%
	{
%
%
\colorlet{mycolor1}{color3p0Hz}
\colorlet{mycolor2}{color4p8Hz}
\colorlet{mycolor3}{color6p9Hz}
\colorlet{mycolor4}{color9p9Hz}
\colorlet{mycolor5}{colorRig}
\begin{tikzpicture}

\begin{axis}[%
width=0.951\figurewidth,
height=\figureheight,
at={(0\figurewidth,0\figureheight)},
scale only axis,
xmin=0.5,
xmax=5.5,
xtick={1,2,3,4,5},
xticklabels={{3.0 Hz},{4.8 Hz},{6.9 Hz},{9.9 Hz},{Rigid}},
xlabel style={font=\color{black}},
xlabel={Condition (-)},
ymin=0,
ymax=6,
ylabel style={font=\color{black}},
ylabel={Velocity Gain (-)},
axis background/.style={fill=white},
axis x line*=bottom,
axis y line*=left
]

\addplot[area legend, line width=2.0pt, draw=black, fill=mycolor1, forget plot]
table[row sep=crcr] {%
x	y\\
0.6	1.71382058763599\\
1.4	1.71382058763599\\
1.4	1.70496946938272\\
1.2	1.8267676227716\\
1.4	1.97229290856691\\
1.4	1.94131497725374\\
0.6	1.94131497725374\\
0.6	1.97229290856691\\
0.8	1.8267676227716\\
0.6	1.70496946938272\\
0.6	1.71382058763599\\
}--cycle;
\addplot [color=black, line width=2.0pt, forget plot]
  table[row sep=crcr]{%
0.8	1.8267676227716\\
1.2	1.8267676227716\\
};
\addplot [color=black, line width=2.0pt, forget plot]
  table[row sep=crcr]{%
1	1.17567734148256\\
1	1.71382058763599\\
};
\addplot [color=black, line width=2.0pt, forget plot]
  table[row sep=crcr]{%
1	1.94131497725374\\
1	2.03226775310834\\
};
\addplot [color=black, line width=2.0pt, forget plot]
  table[row sep=crcr]{%
0.9	1.17567734148256\\
1.1	1.17567734148256\\
};
\addplot [color=black, line width=2.0pt, forget plot]
  table[row sep=crcr]{%
0.9	2.03226775310834\\
1.1	2.03226775310834\\
};

\addplot[area legend, line width=2.0pt, draw=black, fill=mycolor2, forget plot]
table[row sep=crcr] {%
x	y\\
1.6	1.90781848661824\\
2.4	1.90781848661824\\
2.4	1.8718227122635\\
2.2	2.10418939307495\\
2.4	2.3644238407889\\
2.4	2.35834599818443\\
1.6	2.35834599818443\\
1.6	2.3644238407889\\
1.8	2.10418939307495\\
1.6	1.8718227122635\\
1.6	1.90781848661824\\
}--cycle;
\addplot [color=black, line width=2.0pt, forget plot]
  table[row sep=crcr]{%
1.8	2.10418939307495\\
2.2	2.10418939307495\\
};
\addplot [color=black, line width=2.0pt, forget plot]
  table[row sep=crcr]{%
2	1.63869503220032\\
2	1.90781848661824\\
};
\addplot [color=black, line width=2.0pt, forget plot]
  table[row sep=crcr]{%
2	2.35834599818443\\
2	4.03507567810733\\
};
\addplot [color=black, line width=2.0pt, forget plot]
  table[row sep=crcr]{%
1.9	1.63869503220032\\
2.1	1.63869503220032\\
};
\addplot [color=black, line width=2.0pt, forget plot]
  table[row sep=crcr]{%
1.9	4.03507567810733\\
2.1	4.03507567810733\\
};

\addplot[area legend, line width=2.0pt, draw=black, fill=mycolor3, forget plot]
table[row sep=crcr] {%
x	y\\
2.6	1.78078416545533\\
3.4	1.78078416545533\\
3.4	1.73963887979972\\
3.2	2.34021031352429\\
3.4	2.57187488402851\\
3.4	2.52693373790765\\
2.6	2.52693373790765\\
2.6	2.57187488402851\\
2.8	2.34021031352429\\
2.6	1.73963887979972\\
2.6	1.78078416545533\\
}--cycle;
\addplot [color=black, line width=2.0pt, forget plot]
  table[row sep=crcr]{%
2.8	2.34021031352429\\
3.2	2.34021031352429\\
};
\addplot [color=black, line width=2.0pt, forget plot]
  table[row sep=crcr]{%
3	1.30594817469287\\
3	1.78078416545533\\
};
\addplot [color=black, line width=2.0pt, forget plot]
  table[row sep=crcr]{%
3	2.52693373790765\\
3	2.83890952335853\\
};
\addplot [color=black, line width=2.0pt, forget plot]
  table[row sep=crcr]{%
2.9	1.30594817469287\\
3.1	1.30594817469287\\
};
\addplot [color=black, line width=2.0pt, forget plot]
  table[row sep=crcr]{%
2.9	2.83890952335853\\
3.1	2.83890952335853\\
};

\addplot[area legend, line width=2.0pt, draw=black, fill=mycolor4, forget plot]
table[row sep=crcr] {%
x	y\\
3.6	1.33426991512041\\
4.4	1.33426991512041\\
4.4	1.3276064266942\\
4.2	1.67963972477042\\
4.4	1.80807490668551\\
4.4	1.79103558891811\\
3.6	1.79103558891811\\
3.6	1.80807490668551\\
3.8	1.67963972477042\\
3.6	1.3276064266942\\
3.6	1.33426991512041\\
}--cycle;
\addplot [color=black, line width=2.0pt, forget plot]
  table[row sep=crcr]{%
3.8	1.67963972477042\\
4.2	1.67963972477042\\
};
\addplot [color=black, line width=2.0pt, forget plot]
  table[row sep=crcr]{%
4	1.14265757689456\\
4	1.33426991512041\\
};
\addplot [color=black, line width=2.0pt, forget plot]
  table[row sep=crcr]{%
4	1.79103558891811\\
4	3.65371137970498\\
};
\addplot [color=black, line width=2.0pt, forget plot]
  table[row sep=crcr]{%
3.9	1.14265757689456\\
4.1	1.14265757689456\\
};
\addplot [color=black, line width=2.0pt, forget plot]
  table[row sep=crcr]{%
3.9	3.65371137970498\\
4.1	3.65371137970498\\
};

\addplot[area legend, line width=2.0pt, draw=black, fill=mycolor5, forget plot]
table[row sep=crcr] {%
x	y\\
4.6	1.00760588174868\\
5.4	1.00760588174868\\
5.4	1.00691711936978\\
5.2	1.01783394504236\\
5.4	1.02264043955763\\
5.4	1.02232721800175\\
4.6	1.02232721800175\\
4.6	1.02264043955763\\
4.8	1.01783394504236\\
4.6	1.00691711936978\\
4.6	1.00760588174868\\
}--cycle;
\addplot [color=black, line width=2.0pt, forget plot]
  table[row sep=crcr]{%
4.8	1.01783394504236\\
5.2	1.01783394504236\\
};
\addplot [color=black, line width=2.0pt, forget plot]
  table[row sep=crcr]{%
5	0.993298457360612\\
5	1.00760588174868\\
};
\addplot [color=black, line width=2.0pt, forget plot]
  table[row sep=crcr]{%
5	1.02232721800175\\
5	1.04357768098857\\
};
\addplot [color=black, line width=2.0pt, forget plot]
  table[row sep=crcr]{%
4.9	0.993298457360612\\
5.1	0.993298457360612\\
};
\addplot [color=black, line width=2.0pt, forget plot]
  table[row sep=crcr]{%
4.9	1.04357768098857\\
5.1	1.04357768098857\\
};
\addplot [color=black, forget plot]
  table[row sep=crcr]{%
0.5	0\\
5.5	0\\
};
\draw[line width=0.5mm] (axis cs:1,0.3) -- (axis cs:1,0.2) -- (axis cs:5,0.2) -- (axis cs:5,0.3);
\draw[line width=0.5mm] (axis cs:2,0.5) -- (axis cs:2,0.4) -- (axis cs:5,0.4) -- (axis cs:5,0.5);
\draw[line width=0.5mm] (axis cs:3,0.7) -- (axis cs:3,0.6) -- (axis cs:5,0.6) -- (axis cs:5,0.7);
\draw[line width=0.5mm] (axis cs:4,0.9) -- (axis cs:4,0.8) -- (axis cs:5,0.8) -- (axis cs:5,0.9);
\draw[line width=0.5mm] (axis cs:1,0.9) -- (axis cs:1,0.8) -- (axis cs:2,0.8) -- (axis cs:2,0.9);
\draw[line width=0.5mm] (axis cs:2,1.1) -- (axis cs:2,1) -- (axis cs:4,1) -- (axis cs:4,1.1);
\node (ref) at (axis cs:2,4.4) {\scalebox{1}{Ref.}};
\end{axis}
\end{tikzpicture}%
\label{fig:stats-gain-all-exp1-friedman}}
	\hfill
	\subfloat[Input Frequency]%
	{
%
%
\colorlet{mycolor1}{color3p0Hz}
\colorlet{mycolor2}{color4p8Hz}
\colorlet{mycolor3}{color6p9Hz}
\colorlet{mycolor4}{color9p9Hz}
\colorlet{mycolor5}{colorRig}
\begin{tikzpicture}

\begin{axis}[%
width=0.951\figurewidth,
height=\figureheight,
at={(0\figurewidth,0\figureheight)},
scale only axis,
xmin=0.5,
xmax=5.5,
xtick={1,2,3,4,5},
xticklabels={{3.0 Hz},{4.8 Hz},{6.9 Hz},{9.9 Hz},{Rigid}},
xlabel style={font=\color{black}},
xlabel={Condition (-)},
ymin=0,
ymax=10,
ylabel style={font=\color{black}},
ylabel={Excitation Frequency (Hz)},
axis background/.style={fill=white},
axis x line*=bottom,
axis y line*=left
]

\addplot[area legend, line width=2.0pt, draw=black, fill=mycolor1, forget plot]
table[row sep=crcr] {%
x	y\\
0.6	2.6368963451019\\
1.4	2.6368963451019\\
1.4	2.62474424244094\\
1.2	3.06544416358625\\
1.4	3.30065359477123\\
1.4	3.28431372549021\\
0.6	3.28431372549021\\
0.6	3.30065359477123\\
0.8	3.06544416358625\\
0.6	2.62474424244094\\
0.6	2.6368963451019\\
}--cycle;
\addplot [color=black, line width=2.0pt, forget plot]
  table[row sep=crcr]{%
0.8	3.06544416358625\\
1.2	3.06544416358625\\
};
\addplot [color=black, line width=2.0pt, forget plot]
  table[row sep=crcr]{%
1	1.76433247865374\\
1	2.6368963451019\\
};
\addplot [color=black, line width=2.0pt, forget plot]
  table[row sep=crcr]{%
1	3.28431372549021\\
1	3.76649170095047\\
};
\addplot [color=black, line width=2.0pt, forget plot]
  table[row sep=crcr]{%
0.9	1.76433247865374\\
1.1	1.76433247865374\\
};
\addplot [color=black, line width=2.0pt, forget plot]
  table[row sep=crcr]{%
0.9	3.76649170095047\\
1.1	3.76649170095047\\
};

\addplot[area legend, line width=2.0pt, draw=black, fill=mycolor2, forget plot]
table[row sep=crcr] {%
x	y\\
1.6	3.85279509895054\\
2.4	3.85279509895054\\
2.4	3.66477387137044\\
2.2	4.50152869330003\\
2.4	4.98756218905473\\
2.4	4.9818975532321\\
1.6	4.9818975532321\\
1.6	4.98756218905473\\
1.8	4.50152869330003\\
1.6	3.66477387137044\\
1.6	3.85279509895054\\
}--cycle;
\addplot [color=black, line width=2.0pt, forget plot]
  table[row sep=crcr]{%
1.8	4.50152869330003\\
2.2	4.50152869330003\\
};
\addplot [color=black, line width=2.0pt, forget plot]
  table[row sep=crcr]{%
2	3.54627766599595\\
2	3.85279509895054\\
};
\addplot [color=black, line width=2.0pt, forget plot]
  table[row sep=crcr]{%
2	4.9818975532321\\
2	5.14142215173153\\
};
\addplot [color=black, line width=2.0pt, forget plot]
  table[row sep=crcr]{%
1.9	3.54627766599595\\
2.1	3.54627766599595\\
};
\addplot [color=black, line width=2.0pt, forget plot]
  table[row sep=crcr]{%
1.9	5.14142215173153\\
2.1	5.14142215173153\\
};

\addplot[area legend, line width=2.0pt, draw=black, fill=mycolor3, forget plot]
table[row sep=crcr] {%
x	y\\
2.6	4.17120372910805\\
3.4	4.17120372910805\\
3.4	4.1325136612021\\
3.2	4.68407196138658\\
3.4	5.64971751412425\\
3.4	5.6378150193615\\
2.6	5.6378150193615\\
2.6	5.64971751412425\\
2.8	4.68407196138658\\
2.6	4.1325136612021\\
2.6	4.17120372910805\\
}--cycle;
\addplot [color=black, line width=2.0pt, forget plot]
  table[row sep=crcr]{%
2.8	4.68407196138658\\
3.2	4.68407196138658\\
};
\addplot [color=black, line width=2.0pt, forget plot]
  table[row sep=crcr]{%
3	2.38476931123085\\
3	4.17120372910805\\
};
\addplot [color=black, line width=2.0pt, forget plot]
  table[row sep=crcr]{%
3	5.6378150193615\\
3	7.27282344229368\\
};
\addplot [color=black, line width=2.0pt, forget plot]
  table[row sep=crcr]{%
2.9	2.38476931123085\\
3.1	2.38476931123085\\
};
\addplot [color=black, line width=2.0pt, forget plot]
  table[row sep=crcr]{%
2.9	7.27282344229368\\
3.1	7.27282344229368\\
};

\addplot[area legend, line width=2.0pt, draw=black, fill=mycolor4, forget plot]
table[row sep=crcr] {%
x	y\\
3.6	4.53021410042001\\
4.4	4.53021410042001\\
4.4	4.51487973909494\\
4.2	5.48228101210914\\
4.4	5.95322466335922\\
4.4	5.85017555007037\\
3.6	5.85017555007037\\
3.6	5.95322466335922\\
3.8	5.48228101210914\\
3.6	4.51487973909494\\
3.6	4.53021410042001\\
}--cycle;
\addplot [color=black, line width=2.0pt, forget plot]
  table[row sep=crcr]{%
3.8	5.48228101210914\\
4.2	5.48228101210914\\
};
\addplot [color=black, line width=2.0pt, forget plot]
  table[row sep=crcr]{%
4	3.8098693759071\\
4	4.53021410042001\\
};
\addplot [color=black, line width=2.0pt, forget plot]
  table[row sep=crcr]{%
4	5.85017555007037\\
4	6.68963665652376\\
};
\addplot [color=black, line width=2.0pt, forget plot]
  table[row sep=crcr]{%
3.9	3.8098693759071\\
4.1	3.8098693759071\\
};
\addplot [color=black, line width=2.0pt, forget plot]
  table[row sep=crcr]{%
3.9	6.68963665652376\\
4.1	6.68963665652376\\
};

\addplot[area legend, line width=2.0pt, draw=black, fill=mycolor5, forget plot]
table[row sep=crcr] {%
x	y\\
4.6	3.91678849456252\\
5.4	3.91678849456252\\
5.4	3.75711949472734\\
5.2	5.60756471475237\\
5.4	5.93477035784772\\
5.4	5.9347703578476\\
4.6	5.9347703578476\\
4.6	5.93477035784772\\
4.8	5.60756471475237\\
4.6	3.75711949472734\\
4.6	3.91678849456252\\
}--cycle;
\addplot [color=black, line width=2.0pt, forget plot]
  table[row sep=crcr]{%
4.8	5.60756471475237\\
5.2	5.60756471475237\\
};
\addplot [color=black, line width=2.0pt, forget plot]
  table[row sep=crcr]{%
5	2.64223385689356\\
5	3.91678849456252\\
};
\addplot [color=black, line width=2.0pt, forget plot]
  table[row sep=crcr]{%
5	5.9347703578476\\
5	6.25024415016211\\
};
\addplot [color=black, line width=2.0pt, forget plot]
  table[row sep=crcr]{%
4.9	2.64223385689356\\
5.1	2.64223385689356\\
};
\addplot [color=black, line width=2.0pt, forget plot]
  table[row sep=crcr]{%
4.9	6.25024415016211\\
5.1	6.25024415016211\\
};
\addplot [color=black, forget plot]
  table[row sep=crcr]{%
0.5	0\\
5.5	0\\
};
\draw[line width=0.2mm,dashed] (axis cs:0.5,3) -- (axis cs:1.5,3);
\draw[line width=0.2mm,dashed] (axis cs:1.5,4.8) -- (axis cs:2.5,4.8);
\draw[line width=0.2mm,dashed] (axis cs:2.5,6.9) -- (axis cs:3.5,6.9);
\draw[line width=0.2mm,dashed] (axis cs:3.5,9.9) -- (axis cs:4.5,9.9);
\draw[line width=0.5mm] (axis cs:2,1.7) -- (axis cs:2,1.5) -- (axis cs:4,1.5) -- (axis cs:4,1.7);
\draw[line width=0.5mm] (axis cs:1,1.4) -- (axis cs:1,1.2) -- (axis cs:2,1.2) -- (axis cs:2,1.4);
\draw[line width=0.5mm] (axis cs:1,1.1) -- (axis cs:1,0.9) -- (axis cs:3,0.9) -- (axis cs:3,1.1);
\draw[line width=0.5mm] (axis cs:1,0.8) -- (axis cs:1,0.6) -- (axis cs:4,0.6) -- (axis cs:4,0.8);
\draw[line width=0.5mm] (axis cs:1,0.5) -- (axis cs:1,0.3) -- (axis cs:5,0.3) -- (axis cs:5,0.5);
\node (ref) at (axis cs:2,5.7) {\scalebox{1}{Ref.}};
\node[anchor=west] (errlabel) at (axis cs:0.51,9.6) {\scalebox{1}{Relative Error:}};
\node (err1) at (axis cs:1,8.7) {\scalebox{1}{$+2.3\%$}};
\node (err2) at (axis cs:2,8.7) {\scalebox{1}{$-6.3\%$}};
\node (err3) at (axis cs:3,8.7) {\scalebox{1}{$-32\%$}};
\node (err4) at (axis cs:4,8.7) {\scalebox{1}{$-45\%$}};
\node (err5) at (axis cs:5,8.7) {\scalebox{1}{n.a.}};
\end{axis}
\end{tikzpicture}%
\label{fig:stats-efreq-all-exp1-friedman}}
	}
	\caption{Box plots comparing the conditions of \ExpOne through the dependent measures: peak hammer velocity \bvalue{\vhat}, gain \bvalue{\Ggain} and excitation frequency \bvalue{\ffrequency}.
	The lower and upper limits of the boxes represent the first and third quartiles, respectively.
	The notches represent the \SI{95}{\percent} confidence interval of the median, and the whiskers show the minimum and maximum values.
	Brackets show relevant statistically significant differences between conditions.
	The condition marked ``Ref.'' was used as reference for \ExpTwo.
	The dashed lines in Fig.~\ref{fig:stats-efreq-all-exp1-friedman} show the resonance frequencies of the conditions.
	The numbers on top show the relative error in percent per condition of the median excitation frequency compared to the resonance frequency. It was calculated as the dimensionless ratio between the difference of excitation frequency and resonance frequency divided by the resonance frequency.
	}
	\label{fig:stats-all-exp1-friedman}
\end{figure*}
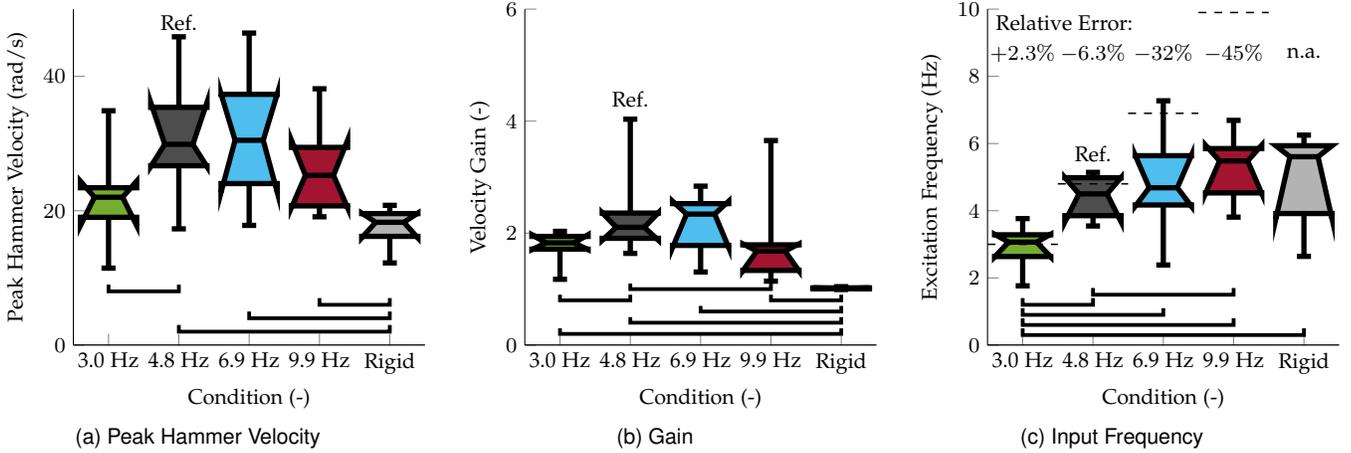

The raw measurement data are available in the archives of the 4TU Centre for Research Data \cite{Aiple2018HumanStiffnesses}.
Fig.~\ref{fig:stats-all-exp1-friedman} shows the box plots of the dependent measures for the five conditions of \ExpOne: peak hammer velocity \bvalue{\vhat}, gain \bvalue{\Ggain}, and excitation frequency \bvalue{\ffrequency}.
Two participants could not finish all conditions because of technical problems with the setup (participant 3 did not finish the \SI{9.9}{\Hz} condition or the rigid condition, and participant 6 did not finish the \SI{3.0}{\Hz} condition or the \SI{4.8}{\Hz} condition). Therefore, the number of samples is 11 for the Friedman's tests and 11 or 12 for the Wilcoxon's signed-rank tests depending on which conditions are compared.

\subsubsection{Peak Hammer Velocity}
The resonance frequency had a significant effect on the peak hammer velocity \vhat, as shown in Fig.~\ref{fig:stats-hammer-velocity-all-exp1-friedman} (\bvalue{\vhat}:
\begin{math}\chisquarevar = \num{27.20}\end{math},
\begin{math}\pvar < \num{0.001}\end{math},
\begin{math}\nvar = \num{11}\end{math}). 

The peak hammer velocity \bvalue{\vhat} was higher for the flexible conditions with \SIlist{4.8;6.9;9.9}{\Hz} resonance frequency than for the rigid condition,
and higher for a \SI{4.8}{\Hz} resonance frequency than for \SI{3.0}{\Hz}
(\SI{4.8}{\Hz} and rigid condition:
\begin{math}\Delta\bvalue{\vhat} = \SI{14.34}{\radian\per\s}\end{math},
\begin{math}\CI = [11.30, 15.43]~\si{\radian\per\s}\end{math},
\begin{math}p = 0.001\end{math},
\begin{math}n = 11\end{math};
\SI{6.9}{\Hz} and rigid condition:
\begin{math}\Delta\bvalue{\vhat} = \SI{9.88}{\radian\per\s} \end{math},
\begin{math}\CI = [8.00 , 16.72]~\si{\radian\per\s}\end{math},
\begin{math}p = 0.001\end{math},
\begin{math}n = 11\end{math};
\SI{9.9}{\Hz} and rigid condition:
\begin{math}\Delta\bvalue{\vhat} = \SI{7.95}{\radian\per\s} \end{math},
\begin{math}\CI = [6.14, 12.39]~\si{\radian\per\s}\end{math},
\begin{math}p = 0.001\end{math},
\begin{math}n = 11\end{math};
\SI{4.8}{\Hz} and \SI{3.0}{\Hz} condition:
\begin{math}\Delta\bvalue{\vhat} = \SI{9.54}{\radian\per\s}\end{math},
\begin{math}\CI = [5.09, 12.67]~\si{\radian\per\s}\end{math},
\begin{math}p = 0.001\end{math},
\begin{math}n = 12\end{math}).

The estimated minimum effect size for \bvalue{\vhat} is \SI{7.8}{\radian\per\s}
(\begin{math}\sigmastdest = \SI{10}{\radian\per\s}\end{math},
\begin{math}n = 11\end{math}).

\subsubsection{Gain}
The resonance frequency had a significant effect on the gain, as shown in Fig.~\ref{fig:stats-gain-all-exp1-friedman}
(\bvalue{\Ggain}:
\begin{math}\chisquarevar = \num{30.25}\end{math},
\begin{math}\pvar < \num{0.001}\end{math},
\begin{math}\nvar = \num{11}\end{math}).

The gain \bvalue{\Ggain} was higher for all flexible conditions than for the rigid condition,
and higher for the \SI{4.8}{\Hz} condition than for the \SIlist{3.0;9.9}{\Hz} conditions
(\SI{3.0}{\Hz} and rigid condition:
\begin{math}\Delta\bvalue{\Ggain} = \SI{0.84}{} \end{math},
\begin{math}\CI = [0.70, 0.98]\end{math},
\begin{math}p = 0.001\end{math},
\begin{math}n = 11\end{math};
\SI{4.8}{\Hz} and rigid condition:
\begin{math}\Delta\bvalue{\Ggain} = \SI{1.16}{} \end{math},
\begin{math}\CI = [0.93, 1.36]\end{math},
\begin{math}p = 0.001\end{math},
\begin{math}n = 11\end{math};
\SI{6.9}{\Hz} and rigid condition:
\begin{math}\Delta\bvalue{\Ggain} = \SI{1.31}{} \end{math},
\begin{math}\CI = [0.73, 1.51]\end{math},
\begin{math}p = 0.001\end{math},
\begin{math}n = 11\end{math};
\SI{9.9}{\Hz} and rigid condition:
\begin{math}\Delta\bvalue{\Ggain} = \SI{0.66}{} \end{math},
\begin{math}\CI = [0.35, 0.79]\end{math},
\begin{math}p = 0.001\end{math},
\begin{math}n = 11\end{math};
\SI{4.8}{\Hz} and \SI{3.0}{\Hz} condition:
\begin{math}\Delta\bvalue{\Ggain} = \SI{0.35}{} \end{math},
\begin{math}\CI = [0.03, 0.66]\end{math},
\begin{math}p < 0.001\end{math},
\begin{math}n = 12\end{math};
\SI{4.8}{\Hz} and \SI{9.9}{\Hz} condition:
\begin{math}\Delta\bvalue{\Ggain} = \SI{0.43}{} \end{math},
\begin{math}\CI = [0.08, 0.83]\end{math},
\begin{math}p = 0.007\end{math},
\begin{math}n = 11\end{math}).

The estimated minimum effect size for \bvalue{\Ggain} is \num{0.51}
(\begin{math}\sigmastdest = \num{0.65}\end{math},
\begin{math}n = 11\end{math}).

\subsubsection{Input Frequency}
The resonance frequency had a significant effect on the excitation frequency, as shown in Fig.~\ref{fig:stats-efreq-all-exp1-friedman} (\bvalue{\ffrequency}:
\begin{math}\chisquarevar = \num{26.84}\end{math},
\begin{math}\pvar < \num{0.001}\end{math},
\begin{math}\nvar = \num{11}\end{math}).

The excitation frequency \bvalue{\ffrequency} was higher for the conditions with \SIlist{4.8;6.9;9.9}{\Hz} resonance frequency and the rigid condition than for the condition with \SI{3.0}{\Hz} resonance frequency,
and higher for the \SI{9.9}{\Hz} condition than for the \SI{4.8}{\Hz} condition
(\SI{4.8}{\Hz} and \SI{3.0}{\Hz} conditions:
\begin{math}\Delta\bvalue{\ffrequency} = \SI{1.58}{\Hz} \end{math},
\begin{math}\CI = [0.91, 1.84]~\si{\Hz}\end{math},
\begin{math}p < 0.001\end{math},
\begin{math}n = 12\end{math};
\SI{6.9}{\Hz} and \SI{3.0}{\Hz} conditions:
\begin{math}\Delta\bvalue{\ffrequency} = \SI{1.58}{\Hz} \end{math},
\begin{math}\CI = [1.02, 2.69]~\si{\Hz}\end{math},
\begin{math}p = 0.001\end{math},
\begin{math}n = 12\end{math};
\SI{9.9}{\Hz} and \SI{3.0}{\Hz} conditions:
\begin{math}\Delta\bvalue{\ffrequency} = \SI{2.73}{\Hz} \end{math},
\begin{math}\CI = [1.79, 2.77]~\si{\Hz}\end{math},
\begin{math}p = 0.001\end{math},
\begin{math}n = 11\end{math};
rigid and \SI{3.0}{\Hz} condition:
\begin{math}\Delta\bvalue{\ffrequency} = \SI{2.37}{\Hz} \end{math},
\begin{math}\CI = [1.13, 2.67]~\si{\Hz}\end{math},
\begin{math}p = 0.001\end{math},
\begin{math}n = 11\end{math};
\SI{9.9}{\Hz} and \SI{4.8}{\Hz} condition:
\begin{math}\Delta\bvalue{\ffrequency} = \SI{0.88}{\Hz} \end{math},
\begin{math}\CI = [0.50, 1.11]~\si{\Hz}\end{math},
\begin{math}p = 0.003\end{math},
\begin{math}n = 11\end{math}).

The estimated minimum effect size for \bvalue{\ffrequency} is \SI{0.74}{\Hz}
(\begin{math}\sigmastdest = \SI{0.95}{\Hz}\end{math},
\begin{math}n = 11\end{math}).

\section{\ExpTwo: \mbox{Teleoperated Manipulation}}
\label{sec:experiment-2}
\subsection{Hypotheses}
\ExpTwo (the teleoperated manipulation experiment) used the optimum stiffness configuration with a resonance frequency of \SI{4.8}{\Hz} that was found during \ExpOne (the direct manipulation experiment).
The following hypotheses were formulated for this experiment, based on \ExpOne and pilot studies for \ExpTwo:

\medskip
\noindent \emph{Hypothesis H2}: Humans can exploit the elasticity of a flexible hammer in teleoperated manipulation to maximize the power transfer in a strike, and to achieve a higher output velocity than they can with a rigid hammer.

\smallskip
\noindent \emph{Hypothesis H2.1}: The absence of visual feedback does not influence the hammering task performance.

\smallskip
\noindent \emph{Hypothesis H2.2}: The absence of force feedback decreases the hammering task performance.

\smallskip
\noindent \emph{Hypothesis H2.3}: Communication delay decreases the hammering task performance.

\subsection{Conditions}
\label{sec:experiment2-conditions}
Four conditions with different feedback were compared:
\begin{itemize}
\item[\FFcond] Visual and force feedback without communication delay.
\item[\NVcond] No visual feedback (the participants were blindfolded).
\item[\NFcond] No force feedback; only visual feedback without communication delay.
\item[\DLcond] Visual and force feedback with \SI{40}{\ms} round-trip communication delay (the participants could see the setup, but the communication delay was asymmetric to achieve the same delay for visual and for force feedback (see Section~\ref{sec:teleoperator-tuning})). The time delay corresponded approximately to round-trip delays encountered on long distance Internet connections, e.g., Amsterdam--Barcelona or Los Angeles--Chicago~\cite{GlobalWonderNetwork}.
\end{itemize}
The order of conditions was balanced using Latin squares to block possible learning and carryover effects.

The participants were verbally informed about the ongoing condition and the conditions were clearly distinguishable:
The participants were blindfolded for the \NVcond~condition, they did not feel the impact in the \NFcond~condition, and they felt a delay-induced damping in the \DLcond~condition.

\subsection{Participants}
\label{sec:participants}
The experiment was performed by 32 participants (26 male, 6 female, all right-handed, age 21-34).
The participants were university students and employees with a higher education background.
None of the participants had participated in \ExpOne or had previous experience with the experiment apparatus.
The experiment had been approved by the human research ethics committee of TU Delft and all participants gave written informed consent before participation.

\subsection{Experiment Procedure}
\label{sec:experiment-flow}
In \ExpTwo, the participants were given less training time than in \ExpOne to explore the resonance mechanism themselves, in order to decrease the experiment time per participant.
To compensate for it, the participants were shown a four-minute video (accessible at \cite{Aiple2017a}) before the start of the experiment, describing the goal of the experiment, giving detailed task instructions, and explaining the basics of mechanical resonance.
As there were (arguably small) differences that make the two experiments difficult to compare directly, the data obtained in \ExpOne only serve as an indicative reference for \ExpTwo and no statistical comparisons are made between the experiments.

Before starting the actual hammering task, the participants had a familiarization phase, which lasted approximately two to three minutes until the participants decided to stop.
The familiarization phase was different from the hammering task, as during this time, the suspended mass was removed and the participants were asked to ``wobble'' the handle to get an impression of the movement required to excite the mechanical resonance of the flexible hammer.
The teleoperation system settings used for this familiarization phase corresponded to the settings of the first experimental condition, in order to ensure that the performance during the first block of hammering trials was based only on feedback available with the first condition.

After the familiarization phase, the suspended mass was put in place and the participants conducted 40 hammering trials with each condition.
After each trial, verbal feedback was given to the participants about their performance in terms of gain.
For the verbal feedback, the words ``Okay,'' ``Good,'' ``Very good,'' and ``Excellent'' were used from the worst to the best performance, with the rating adapted according to the participants' individual performance trend.
To this end, the experiment conductor had a display, which was not visible to the participants, showing the measured gain \Ggain immediately after every trial, allowing him to give the feedback to the participant.
Between two conditions, there was a break to download the data (approx.~\SI{2}{\minute}).
Fig.~\ref{fig:conditions-exp2} visualizes the experiment procedure.
For the analysis, the familiarization phase was not taken into account; only the hammering trials were.

\begin{figure}[!t]
	\footnotesize
\begin{tikzpicture}
[
box/.style={draw},
fambox/.style={box,anchor=west,minimum width=1.2cm,minimum height=1cm,align=center},
condbox/.style={box,anchor=west,minimum height=1cm,minimum width=1.6cm,align=center}
]
\node[fambox] (famil) at (0.4cm,0cm) {Famil.\\2-3 min};
\node[condbox] (cond1) at (1.6cm,0cm){Condition\\40 trials};
\foreach \i in {2,...,4}
{
\node[condbox] (cond\i) at (\i*1.6cm,0cm){\dots};
}
\draw[->,>=latex] (cond1.south) -- ($(cond1.south)+(0cm,-0.8cm)$);
\node[fill=white] (regr) at ($(cond1.south)+(0cm,-0.35cm)$) {10 Highest Gain Trials};
\begin{scope}[shift={($(cond1.south)+(0cm,-0.8cm)$)}]
\node[fill=white] (b) at (0cm,-1em) {$\bvalue{x} = \text{median}(x)$};
\end{scope}
\draw[->,>=latex] ($(famil.north west)+(0cm,0.3cm)$) -- (famil.north west);
\node [anchor=south] (video) at ($(famil.north west)+(0cm,0.3cm)$) {Instructions};
\foreach \i in {2,...,4}
{
\draw[->,>=latex] ($(cond\i.north west)+(0cm,0.3cm)$) -- (cond\i.north west);
\node [anchor=south] (break\i) at ($(cond\i.north west)+(0cm,0.3cm)$) {Break};
}

\end{tikzpicture}
	\caption{Experiment procedure \ExpTwo}
	\label{fig:conditions-exp2}
\end{figure}

\subsection{Teleoperation System Tuning and Identification}
\label{sec:teleoperator-tuning}
\begin{figure}[!t]
	\centering
	{
	\begin{tikzpicture}[
sysblock/.style={node distance=1cm,rectangle,draw,minimum width=width("$e^{-s T}$")+0.3cm,minimum height=height("$e^{-s T}$")+0.3cm},
signal/.style={>=Triangle,->},
measure/.style={>=Triangle[open],->},
sum/.style={draw,circle,minimum size=0.3cm}]
\coordinate (sysref) at (0cm,0cm);
\node[sysblock] (delay1) {\Tdelaylap};
\node[sysblock] (C1) [below=2em of delay1]{\Cone};
\node[sysblock] (C2) [left=  of delay1]{\Ctwo};
\node[sysblock] (delay3) [left=of C2]{\Tdelaylap};
\node[sysblock] (C3) [below=2em of delay3]{\Cthree};
\node[sysblock] (C4) [left= of delay3]{\Cfour};
\node (comm) [fit=(C1) (C4) (delay1)]{};
\node (commleft) [fit=(C3) (C4) (delay3)]{};
\node (commright) [fit=(C1) (C2) (delay1)]{};
\node[sysblock] (Cm) [node distance=0.4cm,below=of commright]{\Cm};
\node[sysblock] (Cs) [node distance=0.4cm,above=of commleft]{\Cs};
\node[sysblock] (Zminv) [node distance=0.6cm,below=of Cm]{\Zminv};
\node[sysblock] (Zsinv) [node distance=0.6cm,above=of Cs]{\Zsinv};
\draw let \p1 = (Zminv),\p2 = (C2) in node at (\x2,\y1) [sum] (summaster){};
\draw let \p1 = (Cm),\p2 = (C2) in node at (\x2,\y1) [sum] (summastercontrol){};
\draw let \p1 = (Cs),\p2 = (delay3) in node at (\x2,\y1) [sum] (sumslavecontrol){};
\draw let \p1 = (Zsinv),\p2 = (delay3) in node at (\x2,\y1) [sum] (sumslave){};

\node[sysblock] (Zh) [node distance=0.4cm,below=of Zminv] {\Zh};
\node[sysblock] (Ze) [node distance=0.4cm,above=of Zsinv] {\Ze};
\draw let \p1 = (Zh),\p2 = (delay3) in node (sumhext) at (\x2,\y1) [sum] {};
\draw let \p1 = (Ze),\p2 = (C2) in node (sumeext) at (\x2,\y1) [sum] {};
\draw let \p1 = (summaster),\p2 = (delay3) in coordinate (fh) at (\x2,\y1) ;

\draw let \p1 = (Ze),\p2 = (C2) in coordinate (fe) at (\x2,\y1) ;
\draw let \p1 = (Zh),\p2 = (commleft) in coordinate (fhext) at (\x2,\y1) ;
\draw let \p1 = (Ze),\p2 = (commright) in coordinate (feext) at (\x2,\y1) ;

\junction{(C4)}{(Zsinv)}
\junction{(C4)}{(Cs)}
\junction{(delay1)}{(Zminv)}
\junction{(delay1)}{(Cm)}
\junction{(sumhext)}{(summaster)}
\junction{(fe)}{(sumslave)}

\draw[signal] (C1) -- (delay1);
\draw[signal] (C3) -- (delay3);
\draw[signal] (Cm) -- (summastercontrol);
\draw[signal] (Cs) -- (sumslavecontrol);
\draw[measure] (summaster) -- (Zminv);
\draw[measure] (sumslave) -- (Zsinv);
\draw[signal] (C2) -- (summastercontrol);
\draw[signal] (summastercontrol) -- (summaster) node [midway, left] {\fmc};
\draw[signal] (delay3) -- (sumslavecontrol);
\draw[signal] (sumslavecontrol) -- (sumslave) node [midway, left] {\fsc};
\draw[measure] (Zminv) -| (C1) node [near start, above] {\vm};
\draw[measure] (Zsinv) -| (C4) node [near start, above] {\vs};
\draw[signal] (delay1) |- (sumslavecontrol);
\draw[signal] (C4) |- (summastercontrol);
\draw let \p1 = (Zminv),\p2 = (delay1) in coordinate (Zminvjunction) at (\x2,\y1);
\draw[measure] (Zminvjunction) |- (Zh);
\draw[measure] (Zminvjunction) |- (Cm);
\draw[measure] (Zh) -- (sumhext);
\draw[measure] (fh) -- (summaster);
\draw[measure] (sumhext) -- (C3);
\draw let \p1 = (Zsinv),\p2 = (C4) in coordinate (Zsinvjunction) at (\x2,\y1);
\draw[measure] (Zsinvjunction) |- (Cs);
\draw[measure] (Zsinvjunction) |- (Ze);
\draw[measure] (Ze) -- (sumeext);
\draw[measure] (sumeext) -- (C2);
\draw[measure] (sumeext) |- (sumslave);
\draw[measure] (fhext) -- (sumhext) node [near start, above] {\fhext};
\draw[measure] (feext) -- (sumeext) node [near start, above] {\feext};

\node (plus) [node distance=1pt,above left=of summaster.west]{$+$};
\node (plus) [node distance=1pt,above right=of summaster.north]{$+$};
\node (plus) [node distance=1pt,above left=of summastercontrol.west]{$-$};
\node (plus) [node distance=1pt,above right=of summastercontrol.north]{$-$};
\node (plus) [node distance=1pt,below right=1pt and -2pt of summastercontrol.east]{$-$};

\node (plus) [node distance=1pt,above right=of sumslave.east]{$-$};
\node (plus) [node distance=1pt,below right=of sumslave.south]{$+$};
\node (plus) [node distance=1pt,below right=of sumslavecontrol.east]{$+$};
\node (plus) [node distance=1pt,above left=1pt and -2pt of sumslavecontrol.west]{$-$};
\node (plus) [node distance=1pt,below left=1pt and 0pt of sumslavecontrol.south]{$+$};
\node (plus) [node distance=1pt,above left=of sumhext.west]{$+$};
\node (plus) [node distance=1pt,above right=of sumhext.east]{$-$};
\node (plus) [node distance=1pt,above left=of sumeext.west]{$+$};
\node (plus) [node distance=1pt,above right=of sumeext.east]{$+$};

\end{tikzpicture}
	}
	\caption{Controller architecture of the teleoperator.
    Three different settings were used for the controller gains \Cm, \Cs, \Cone, \Ctwo, \Cthree, \Cfour and the delay \Tdelay for the conditions of \ExpTwo.
    Note that communication delay was implemented only on the forward channels (handle device to tool device) and not on the feedback channels (tool device to handle device) in order to achieve communication delay for the visual feedback.
    Therefore, \Tdelay represents the round-trip time and not the one-way delay time.
    \Ze is the environment impedance; \Zh is the impedance of the human operator; \Zm is the impedance of the handle device; \Zs is the impedance of the tool device; \fe is the sum of external forces on the hammer; \fh is the force intentionally exerted by the human operator on the handle device; \fmc and \fsc are the forces commanded by the controller to the handle device and tool device; and \vm and \vs are the velocities of the handle device and the tool device.
    (Adapted from \cite{Hashtrudi-zaad2001})}
	\label{fig:controller-architecture}
\end{figure}
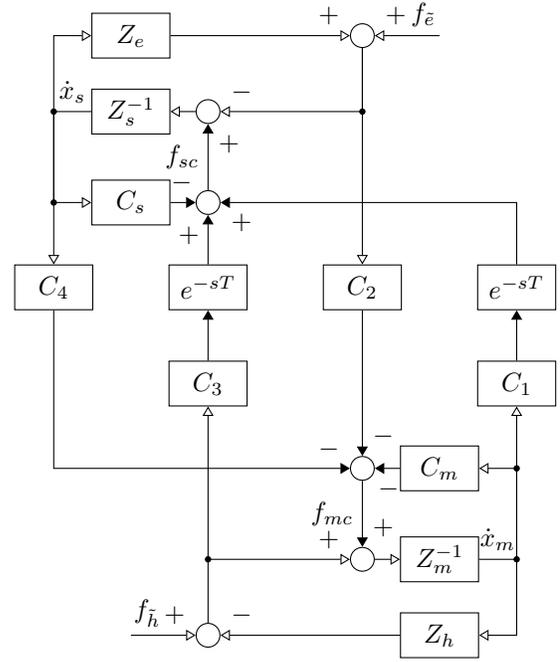

Different controllers were programmed in the teleoperator control system to provide force feedback without communication delay, force feedback with communication delay, or no force feedback.
The structure of the controllers was based on the Lawrence architecture~\cite{Lawrence1993} (see~Fig.~\ref{fig:controller-architecture}), using different tuning gains \Cm, \Cs, \Cone, \Ctwo, \Cthree, \Cfour.
These three controllers covered the four experiment conditions (see Table~\ref{tbl:controller-gains}).

Controller~1 was tuned for optimal transparency according to the rules of Hashtrudi-Zaad and Salcudean~\cite{Hashtrudi-zaad2001}.
Controller~2 was derived from Controller~1 by adding a communication delay of $\Tdelay = \SI{40}{\ms}$ only in the tool device to handle device direction.
This way, a visual communication delay identical to the force feedback communication delay could be simulated without the need to install a sophisticated video transmission system.
Controller~3 was derived from Controller~1 by disabling the force feedback.

\begin{table}[!t]
	\centering
	\begin{threeparttable}
		\caption{Controller gains of the system shown in~Fig.~\ref{fig:controller-architecture}}
		\label{tbl:controller-gains}
		\renewcommand*{\arraystretch}{1.6}
		\begin{tabular}{p{5ex}p{6ex}ccccc}
			\toprule
			Contr. & Cond. & \multicolumn{5}{c}{Parameter}\\
			& & \Cm = - \Cfour & \Cs = \Cone  & \Ctwo & \Cthree & \Tdelay \\
			\midrule
			1 & \FFcond, \NVcond    & 0.8 + $\frac{10}{\slap}$ & 0.8 + $\frac{10}{\slap}$ & 1 & 1 & \SI{0}{\s} \\
			2 & \DLcond  & 0.8 + $\frac{10}{\slap}$ & 0.8 + $\frac{10}{\slap}$ & 1 & 1 & \SI{40}{\ms}\\
			3 & \NFcond &                        0 & 0.8 + $\frac{10}{\slap}$ & 0 & 1 & \SI{0}{\s}\\
			\bottomrule
		\end{tabular}
	\end{threeparttable}
\end{table}

A system identification was carried out to assess the performance of the teleoperator.
A sine sweep identification input signal (\SIrange{0.1}{20}{\Hz} over \SI{100}{\s}) was used to analyze the teleoperator transparency by comparing the transmitted impedance \Zto for the three controllers with the environment impedance \Ze.
The environment impedance is the impedance of the flexible hammer in free air.
The transmitted impedance \Zto should be close to the environment impedance \Ze in order to make the teleoperated hammer behave like the directly manipulated hammer in \ExpOne.
Only the free air behavior of the setup was measured as this was the relevant mode of operation for this experiment.

\begin{figure}[!t]
	\centering
	\setlength{\figurewidth}{.8\columnwidth}
	\setlength{\figureheight}{\plotheightB}
	\input{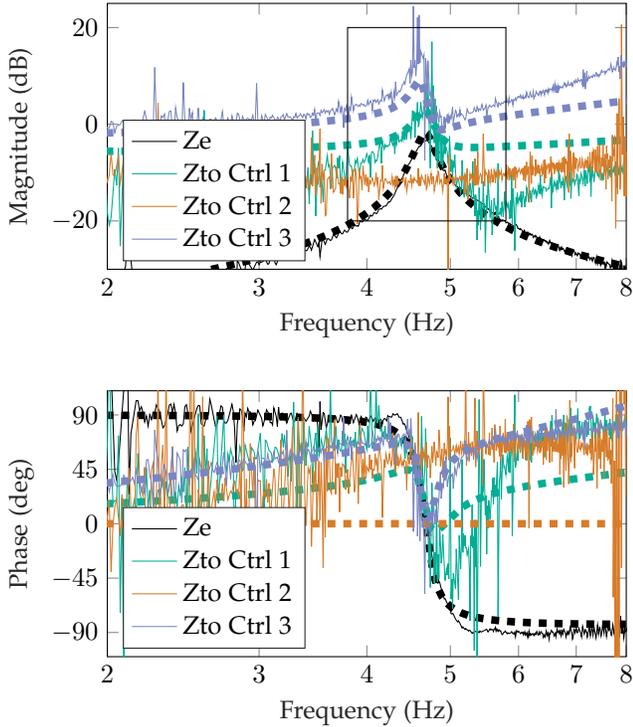}
	\caption{Bode plots of the modeled and measured environment impedance \Ze and modeled and measured transmitted impedance \Zto for the three controllers.
		The box in the magnitude plots from \SIrange{3.8}{5.8}{\Hz} frequency and from \SIrange{0}{20}{\dB} magnitude highlights the critical region around the resonance frequency of \SI{4.8}{\Hz}.
	}
	\label{fig:ztoplot-all}
\end{figure}

Fig.~\ref{fig:ztoplot-all} shows the bode plots of the impedance \Zto transmitted through the teleoperator compared to the environment impedance \Ze for the three controllers. 

The black dashed line shows the modeled curve of \Ze:
\begin{dmath}
\Ze = \Kvar\,\frac{\kstiffness\;(\mmass\,\slap + b)}{\mmass\,\slap^2+\bdamping\,\slap+\kstiffness}
\end{dmath},
with the numeric values obtained from the system identification measurement
(\begin{math}\Kvar = \num{0.5}\end{math},
\begin{math}\mmass = \SI{2.59}{\g\m\squared}\end{math}, 
\begin{math}\kstiffness = \SI{2.23}{\N\m\per\radian}\end{math},
\begin{math}\bdamping = \SI[prefixes-as-symbols=false]{3e-3}{\N\m\s\per\radian}\end{math}).

The other dashed lines show the modeled curves of \Zto for the three controllers with the parameters in Table~\ref{tbl:controller-gains}.
They were derived from the \Ze model using the Haptic Analysis Toolbox~\cite{Christiansson2007}.

Controller~1 with optimal transparency gains has a good fidelity of the transmitted impedance.
The frequency of the maximum impedance magnitude of the transmitted impedance closely matches the environment impedance, and the magnitude profiles of both curves are generally close within the boxed region.
The maximum impedance is at least \SI{16}{\dB} higher than at the frequencies \SI{1}{\Hz} higher or lower in both curves.
This can be compared to a signal-to-noise ratio for the sensing of the mechanical resonance.
The higher the signal-to-noise ratio, the easier it should be for the operator to sense and excite the mechanical resonance with only impedance information as feedback.

The impedance maximum is slightly shifted for Controller~2 and is at approximately \SI{4.6}{\Hz} instead of \SI{4.8}{\Hz}.
Also the ratio between the minimum magnitude and the magnitude \SI{1}{\Hz} higher or lower is only about \SI{5.5}{\dB}.
Thus, with Controller~2 it should be more difficult for the operator to sense and excite the mechanical resonance with only impedance information as feedback than is the case with Controller~1 .

For Controller~3, no impedance minimum can be distinguished and therefore the mechanical resonance cannot be identified on the basis of impedance feedback.

\subsection{Results}
\begin{figure*}[!t]
	{
	\footnotesize
	\setlength{\figurewidth}{.27\textwidth}
	\setlength{\figureheight}{\plotheightA}
	\subfloat[Peak Hammer Velocity]%
	{
%
%
\colorlet{mycolor1}{color4p8Hz}
\colorlet{mycolor2}{colorFF}
\colorlet{mycolor3}{colorNV}
\colorlet{mycolor4}{colorNF}
\colorlet{mycolor5}{colorDL}
\begin{tikzpicture}

\begin{axis}[%
width=0.951\figurewidth,
height=\figureheight,
at={(0\figurewidth,0\figureheight)},
scale only axis,
xmin=0.5,
xmax=5.5,
xtick={1,2,3,4,5},
xticklabels={{Ref.},{FF},{NV},{NF},{DL}},
xlabel style={font=\color{black}},
xlabel={Condition (-)},
ymin=0,
ymax=50,
ylabel style={font=\color{black}},
ylabel={Peak Hammer Velocity (rad/s)},
axis background/.style={fill=white},
axis x line*=bottom,
axis y line*=left
]

\addplot[area legend, line width=2.0pt, draw=black, fill=mycolor1, forget plot]
table[row sep=crcr] {%
x	y\\
0.6	26.693859299537\\
1.4	26.693859299537\\
1.4	26.5768412015772\\
1.2	29.8853051935359\\
1.4	36.1106103836636\\
1.4	35.3884466096858\\
0.6	35.3884466096858\\
0.6	36.1106103836636\\
0.8	29.8853051935359\\
0.6	26.5768412015772\\
0.6	26.693859299537\\
}--cycle;
\addplot [color=black, line width=2.0pt, forget plot]
  table[row sep=crcr]{%
0.800000000000001	29.8853051935359\\
1.2	29.8853051935359\\
};
\addplot [color=black, line width=2.0pt, forget plot]
  table[row sep=crcr]{%
1	17.3049777442583\\
1	26.693859299537\\
};
\addplot [color=black, line width=2.0pt, forget plot]
  table[row sep=crcr]{%
1	35.3884466096858\\
1	45.8902789022242\\
};
\addplot [color=black, line width=2.0pt, forget plot]
  table[row sep=crcr]{%
0.899999999999999	17.3049777442583\\
1.1	17.3049777442583\\
};
\addplot [color=black, line width=2.0pt, forget plot]
  table[row sep=crcr]{%
0.899999999999999	45.8902789022242\\
1.1	45.8902789022242\\
};

\addplot[area legend, line width=2.0pt, draw=black, fill=mycolor2, forget plot]
table[row sep=crcr] {%
x	y\\
1.6	18.52703143389\\
2.4	18.52703143389\\
2.4	19.4375573021117\\
2.2	23.1765502633206\\
2.4	27.2955219626474\\
2.4	28.5885210957921\\
1.6	28.5885210957921\\
1.6	27.2955219626474\\
1.8	23.1765502633206\\
1.6	19.4375573021117\\
1.6	18.52703143389\\
}--cycle;
\addplot [color=black, line width=2.0pt, forget plot]
  table[row sep=crcr]{%
1.8	23.1765502633206\\
2.2	23.1765502633206\\
};
\addplot [color=black, line width=2.0pt, forget plot]
  table[row sep=crcr]{%
2	10.3220059723399\\
2	18.52703143389\\
};
\addplot [color=black, line width=2.0pt, forget plot]
  table[row sep=crcr]{%
2	28.5885210957921\\
2	37.7420003394766\\
};
\addplot [color=black, line width=2.0pt, forget plot]
  table[row sep=crcr]{%
1.9	10.3220059723399\\
2.1	10.3220059723399\\
};
\addplot [color=black, line width=2.0pt, forget plot]
  table[row sep=crcr]{%
1.9	37.7420003394766\\
2.1	37.7420003394766\\
};

\addplot[area legend, line width=2.0pt, draw=black, fill=mycolor3, forget plot]
table[row sep=crcr] {%
x	y\\
2.6	18.1419438964896\\
3.4	18.1419438964896\\
3.4	18.8843880400285\\
3.2	24.3151034817051\\
3.4	27.3587672073621\\
3.4	28.9800886464596\\
2.6	28.9800886464596\\
2.6	27.3587672073621\\
2.8	24.3151034817051\\
2.6	18.8843880400285\\
2.6	18.1419438964896\\
}--cycle;
\addplot [color=black, line width=2.0pt, forget plot]
  table[row sep=crcr]{%
2.8	24.3151034817051\\
3.2	24.3151034817051\\
};
\addplot [color=black, line width=2.0pt, forget plot]
  table[row sep=crcr]{%
3	8.7436069669109\\
3	18.1419438964896\\
};
\addplot [color=black, line width=2.0pt, forget plot]
  table[row sep=crcr]{%
3	28.9800886464596\\
3	42.9078552088367\\
};
\addplot [color=black, line width=2.0pt, forget plot]
  table[row sep=crcr]{%
2.9	8.7436069669109\\
3.1	8.7436069669109\\
};
\addplot [color=black, line width=2.0pt, forget plot]
  table[row sep=crcr]{%
2.9	42.9078552088367\\
3.1	42.9078552088367\\
};

\addplot[area legend, line width=2.0pt, draw=black, fill=mycolor4, forget plot]
table[row sep=crcr] {%
x	y\\
3.6	18.6440493100342\\
4.4	18.6440493100342\\
4.4	20.5151356171182\\
4.2	22.6177178785562\\
4.4	25.9449509164531\\
4.4	27.3533485228917\\
3.6	27.3533485228917\\
3.6	25.9449509164531\\
3.8	22.6177178785562\\
3.6	20.5151356171182\\
3.6	18.6440493100342\\
}--cycle;
\addplot [color=black, line width=2.0pt, forget plot]
  table[row sep=crcr]{%
3.8	22.6177178785562\\
4.2	22.6177178785562\\
};
\addplot [color=black, line width=2.0pt, forget plot]
  table[row sep=crcr]{%
4	11.8110821448587\\
4	18.6440493100342\\
};
\addplot [color=black, line width=2.0pt, forget plot]
  table[row sep=crcr]{%
4	27.3533485228917\\
4	37.9375665279238\\
};
\addplot [color=black, line width=2.0pt, forget plot]
  table[row sep=crcr]{%
3.9	11.8110821448587\\
4.1	11.8110821448587\\
};
\addplot [color=black, line width=2.0pt, forget plot]
  table[row sep=crcr]{%
3.9	37.9375665279238\\
4.1	37.9375665279238\\
};

\addplot[area legend, line width=2.0pt, draw=black, fill=mycolor5, forget plot]
table[row sep=crcr] {%
x	y\\
4.6	18.3153276861628\\
5.4	18.3153276861628\\
5.4	20.0531001864428\\
5.2	23.6769864060684\\
5.4	27.0228355972544\\
5.4	31.0632209380532\\
4.6	31.0632209380532\\
4.6	27.0228355972544\\
4.8	23.6769864060684\\
4.6	20.0531001864428\\
4.6	18.3153276861628\\
}--cycle;
\addplot [color=black, line width=2.0pt, forget plot]
  table[row sep=crcr]{%
4.8	23.6769864060684\\
5.2	23.6769864060684\\
};
\addplot [color=black, line width=2.0pt, forget plot]
  table[row sep=crcr]{%
5	12.5682048318958\\
5	18.3153276861628\\
};
\addplot [color=black, line width=2.0pt, forget plot]
  table[row sep=crcr]{%
5	31.0632209380532\\
5	36.8484667153988\\
};
\addplot [color=black, line width=2.0pt, forget plot]
  table[row sep=crcr]{%
4.9	12.5682048318958\\
5.1	12.5682048318958\\
};
\addplot [color=black, line width=2.0pt, forget plot]
  table[row sep=crcr]{%
4.9	36.8484667153988\\
5.1	36.8484667153988\\
};
\addplot [color=black, forget plot]
  table[row sep=crcr]{%
0.5	0\\
5.5	0\\
};
\end{axis}
\end{tikzpicture}%
\label{fig:stats-hammer-velocity-all-friedman}}
	\hfill
	\subfloat[Gain]%
	{
%
%
\colorlet{mycolor1}{color4p8Hz}
\colorlet{mycolor2}{colorFF}
\colorlet{mycolor3}{colorNV}
\colorlet{mycolor4}{colorNF}
\colorlet{mycolor5}{colorDL}
\begin{tikzpicture}

\begin{axis}[%
width=0.951\figurewidth,
height=\figureheight,
at={(0\figurewidth,0\figureheight)},
scale only axis,
xmin=0.5,
xmax=5.5,
xtick={1,2,3,4,5},
xticklabels={{Ref.},{FF},{NV},{NF},{DL}},
xlabel style={font=\color{black}},
xlabel={Condition (-)},
ymin=0,
ymax=6,
ylabel style={font=\color{black}},
ylabel={Gain (-)},
axis background/.style={fill=white},
axis x line*=bottom,
axis y line*=left
]

\addplot[area legend, line width=2.0pt, draw=black, fill=mycolor1, forget plot]
table[row sep=crcr] {%
x	y\\
0.6	1.90781848661824\\
1.4	1.90781848661824\\
1.4	1.8718227122635\\
1.2	2.10418939307495\\
1.4	2.3644238407889\\
1.4	2.35834599818443\\
0.6	2.35834599818443\\
0.6	2.3644238407889\\
0.8	2.10418939307495\\
0.6	1.8718227122635\\
0.6	1.90781848661824\\
}--cycle;
\addplot [color=black, line width=2.0pt, forget plot]
  table[row sep=crcr]{%
0.8	2.10418939307495\\
1.2	2.10418939307495\\
};
\addplot [color=black, line width=2.0pt, forget plot]
  table[row sep=crcr]{%
1	1.63869503220032\\
1	1.90781848661824\\
};
\addplot [color=black, line width=2.0pt, forget plot]
  table[row sep=crcr]{%
1	2.35834599818443\\
1	4.03507567810733\\
};
\addplot [color=black, line width=2.0pt, forget plot]
  table[row sep=crcr]{%
0.9	1.63869503220032\\
1.1	1.63869503220032\\
};
\addplot [color=black, line width=2.0pt, forget plot]
  table[row sep=crcr]{%
0.9	4.03507567810733\\
1.1	4.03507567810733\\
};

\addplot[area legend, line width=2.0pt, draw=black, fill=mycolor2, forget plot]
table[row sep=crcr] {%
x	y\\
1.6	2.34047880304706\\
2.4	2.34047880304706\\
2.4	2.41736458733725\\
2.2	2.58231165261176\\
2.4	2.85588660335873\\
2.4	2.95252268896772\\
1.6	2.95252268896772\\
1.6	2.85588660335873\\
1.8	2.58231165261176\\
1.6	2.41736458733725\\
1.6	2.34047880304706\\
}--cycle;
\addplot [color=black, line width=2.0pt, forget plot]
  table[row sep=crcr]{%
1.8	2.58231165261176\\
2.2	2.58231165261176\\
};
\addplot [color=black, line width=2.0pt, forget plot]
  table[row sep=crcr]{%
2	1.73893190861546\\
2	2.34047880304706\\
};
\addplot [color=black, line width=2.0pt, forget plot]
  table[row sep=crcr]{%
2	2.95252268896772\\
2	4.59062227183483\\
};
\addplot [color=black, line width=2.0pt, forget plot]
  table[row sep=crcr]{%
1.9	1.73893190861546\\
2.1	1.73893190861546\\
};
\addplot [color=black, line width=2.0pt, forget plot]
  table[row sep=crcr]{%
1.9	4.59062227183483\\
2.1	4.59062227183483\\
};

\addplot[area legend, line width=2.0pt, draw=black, fill=mycolor3, forget plot]
table[row sep=crcr] {%
x	y\\
2.6	2.2924399228255\\
3.4	2.2924399228255\\
3.4	2.38256261565787\\
3.2	2.64969566676826\\
3.4	2.90019382041577\\
3.4	3.05440955308445\\
2.6	3.05440955308445\\
2.6	2.90019382041577\\
2.8	2.64969566676826\\
2.6	2.38256261565787\\
2.6	2.2924399228255\\
}--cycle;
\addplot [color=black, line width=2.0pt, forget plot]
  table[row sep=crcr]{%
2.8	2.64969566676826\\
3.2	2.64969566676826\\
};
\addplot [color=black, line width=2.0pt, forget plot]
  table[row sep=crcr]{%
3	1.6964694123451\\
3	2.2924399228255\\
};
\addplot [color=black, line width=2.0pt, forget plot]
  table[row sep=crcr]{%
3	3.05440955308445\\
3	5.55537840358614\\
};
\addplot [color=black, line width=2.0pt, forget plot]
  table[row sep=crcr]{%
2.9	1.6964694123451\\
3.1	1.6964694123451\\
};
\addplot [color=black, line width=2.0pt, forget plot]
  table[row sep=crcr]{%
2.9	5.55537840358614\\
3.1	5.55537840358614\\
};

\addplot[area legend, line width=2.0pt, draw=black, fill=mycolor4, forget plot]
table[row sep=crcr] {%
x	y\\
3.6	2.35223327537859\\
4.4	2.35223327537859\\
4.4	2.49284883483097\\
4.2	2.80173314446875\\
4.4	2.96064784510851\\
4.4	3.19174307326694\\
3.6	3.19174307326694\\
3.6	2.96064784510851\\
3.8	2.80173314446875\\
3.6	2.49284883483097\\
3.6	2.35223327537859\\
}--cycle;
\addplot [color=black, line width=2.0pt, forget plot]
  table[row sep=crcr]{%
3.8	2.80173314446875\\
4.2	2.80173314446875\\
};
\addplot [color=black, line width=2.0pt, forget plot]
  table[row sep=crcr]{%
4	1.66354028054209\\
4	2.35223327537859\\
};
\addplot [color=black, line width=2.0pt, forget plot]
  table[row sep=crcr]{%
4	3.19174307326694\\
4	4.80998603535332\\
};
\addplot [color=black, line width=2.0pt, forget plot]
  table[row sep=crcr]{%
3.9	1.66354028054209\\
4.1	1.66354028054209\\
};
\addplot [color=black, line width=2.0pt, forget plot]
  table[row sep=crcr]{%
3.9	4.80998603535332\\
4.1	4.80998603535332\\
};

\addplot[area legend, line width=2.0pt, draw=black, fill=mycolor5, forget plot]
table[row sep=crcr] {%
x	y\\
4.6	2.42082200056858\\
5.4	2.42082200056858\\
5.4	2.45398692659626\\
5.2	2.73306701795387\\
5.4	3.00305335267064\\
5.4	3.16896078063108\\
4.6	3.16896078063108\\
4.6	3.00305335267064\\
4.8	2.73306701795387\\
4.6	2.45398692659626\\
4.6	2.42082200056858\\
}--cycle;
\addplot [color=black, line width=2.0pt, forget plot]
  table[row sep=crcr]{%
4.8	2.73306701795387\\
5.2	2.73306701795387\\
};
\addplot [color=black, line width=2.0pt, forget plot]
  table[row sep=crcr]{%
5	1.72465821866161\\
5	2.42082200056858\\
};
\addplot [color=black, line width=2.0pt, forget plot]
  table[row sep=crcr]{%
5	3.16896078063108\\
5	3.52652476049494\\
};
\addplot [color=black, line width=2.0pt, forget plot]
  table[row sep=crcr]{%
4.9	1.72465821866161\\
5.1	1.72465821866161\\
};
\addplot [color=black, line width=2.0pt, forget plot]
  table[row sep=crcr]{%
4.9	3.52652476049494\\
5.1	3.52652476049494\\
};
\addplot [color=black, forget plot]
  table[row sep=crcr]{%
0.5	0\\
5.5	0\\
};
\end{axis}
\end{tikzpicture}%
\label{fig:stats-gain-all-friedman}}
	\hfill
	\subfloat[Input Frequency]%
	{
%
%
\colorlet{mycolor1}{color4p8Hz}
\colorlet{mycolor2}{colorFF}
\colorlet{mycolor3}{colorNV}
\colorlet{mycolor4}{colorNF}
\colorlet{mycolor5}{colorDL}
\begin{tikzpicture}

\begin{axis}[%
width=0.951\figurewidth,
height=\figureheight,
at={(0\figurewidth,0\figureheight)},
scale only axis,
xmin=0.5,
xmax=5.5,
xtick={1,2,3,4,5},
xticklabels={{Ref.},{FF},{NV},{NF},{DL}},
xlabel style={font=\color{black}},
xlabel={Condition (-)},
ymin=0,
ymax=10,
ylabel style={font=\color{black}},
ylabel={Excitation Frequency (Hz)},
axis background/.style={fill=white},
axis x line*=bottom,
axis y line*=left
]

\addplot[area legend, line width=2.0pt, draw=black, fill=mycolor1, forget plot]
table[row sep=crcr] {%
x	y\\
0.6	3.85279509895054\\
1.4	3.85279509895054\\
1.4	3.66477387137044\\
1.2	4.50152869330003\\
1.4	4.98756218905473\\
1.4	4.9818975532321\\
0.6	4.9818975532321\\
0.6	4.98756218905473\\
0.8	4.50152869330003\\
0.6	3.66477387137044\\
0.6	3.85279509895054\\
}--cycle;
\addplot [color=black, line width=2.0pt, forget plot]
  table[row sep=crcr]{%
0.8	4.50152869330003\\
1.2	4.50152869330003\\
};
\addplot [color=black, line width=2.0pt, forget plot]
  table[row sep=crcr]{%
1	3.54627766599595\\
1	3.85279509895054\\
};
\addplot [color=black, line width=2.0pt, forget plot]
  table[row sep=crcr]{%
1	4.9818975532321\\
1	5.14142215173153\\
};
\addplot [color=black, line width=2.0pt, forget plot]
  table[row sep=crcr]{%
0.9	3.54627766599595\\
1.1	3.54627766599595\\
};
\addplot [color=black, line width=2.0pt, forget plot]
  table[row sep=crcr]{%
0.9	5.14142215173153\\
1.1	5.14142215173153\\
};

\addplot[area legend, line width=2.0pt, draw=black, fill=mycolor2, forget plot]
table[row sep=crcr] {%
x	y\\
1.6	3.28837828837832\\
2.4	3.28837828837832\\
2.4	3.51738291874812\\
2.2	4.230452557684\\
2.4	4.60064303240542\\
2.4	4.76212074089431\\
1.6	4.76212074089431\\
1.6	4.60064303240542\\
1.8	4.230452557684\\
1.6	3.51738291874812\\
1.6	3.28837828837832\\
}--cycle;
\addplot [color=black, line width=2.0pt, forget plot]
  table[row sep=crcr]{%
1.8	4.230452557684\\
2.2	4.230452557684\\
};
\addplot [color=black, line width=2.0pt, forget plot]
  table[row sep=crcr]{%
2	2.47526269031652\\
2	3.28837828837832\\
};
\addplot [color=black, line width=2.0pt, forget plot]
  table[row sep=crcr]{%
2	4.76212074089431\\
2	6.66933440042717\\
};
\addplot [color=black, line width=2.0pt, forget plot]
  table[row sep=crcr]{%
1.9	2.47526269031652\\
2.1	2.47526269031652\\
};
\addplot [color=black, line width=2.0pt, forget plot]
  table[row sep=crcr]{%
1.9	6.66933440042717\\
2.1	6.66933440042717\\
};

\addplot[area legend, line width=2.0pt, draw=black, fill=mycolor3, forget plot]
table[row sep=crcr] {%
x	y\\
2.6	3.07575380565865\\
3.4	3.07575380565865\\
3.4	3.43223831390558\\
3.2	3.78273338201642\\
3.4	4.04055245371739\\
3.4	4.1546651523422\\
2.6	4.1546651523422\\
2.6	4.04055245371739\\
2.8	3.78273338201642\\
2.6	3.43223831390558\\
2.6	3.07575380565865\\
}--cycle;
\addplot [color=black, line width=2.0pt, forget plot]
  table[row sep=crcr]{%
2.8	3.78273338201642\\
3.2	3.78273338201642\\
};
\addplot [color=black, line width=2.0pt, forget plot]
  table[row sep=crcr]{%
3	2.53222983671926\\
3	3.07575380565865\\
};
\addplot [color=black, line width=2.0pt, forget plot]
  table[row sep=crcr]{%
3	4.1546651523422\\
3	5.44995544995507\\
};
\addplot [color=black, line width=2.0pt, forget plot]
  table[row sep=crcr]{%
2.9	2.53222983671926\\
3.1	2.53222983671926\\
};
\addplot [color=black, line width=2.0pt, forget plot]
  table[row sep=crcr]{%
2.9	5.44995544995507\\
3.1	5.44995544995507\\
};

\addplot[area legend, line width=2.0pt, draw=black, fill=mycolor4, forget plot]
table[row sep=crcr] {%
x	y\\
3.6	3.35583258725111\\
4.4	3.35583258725111\\
4.4	3.59679710706448\\
4.2	3.8577813514896\\
4.4	4.17552913281508\\
4.4	4.47979556386643\\
3.6	4.47979556386643\\
3.6	4.17552913281508\\
3.8	3.8577813514896\\
3.6	3.59679710706448\\
3.6	3.35583258725111\\
}--cycle;
\addplot [color=black, line width=2.0pt, forget plot]
  table[row sep=crcr]{%
3.8	3.8577813514896\\
4.2	3.8577813514896\\
};
\addplot [color=black, line width=2.0pt, forget plot]
  table[row sep=crcr]{%
4	2.4789580389506\\
4	3.35583258725111\\
};
\addplot [color=black, line width=2.0pt, forget plot]
  table[row sep=crcr]{%
4	4.47979556386643\\
4	5.46513364192945\\
};
\addplot [color=black, line width=2.0pt, forget plot]
  table[row sep=crcr]{%
3.9	2.4789580389506\\
4.1	2.4789580389506\\
};
\addplot [color=black, line width=2.0pt, forget plot]
  table[row sep=crcr]{%
3.9	5.46513364192945\\
4.1	5.46513364192945\\
};

\addplot[area legend, line width=2.0pt, draw=black, fill=mycolor5, forget plot]
table[row sep=crcr] {%
x	y\\
4.6	3.56734377850382\\
5.4	3.56734377850382\\
5.4	3.70986136110682\\
5.2	4.07621381886077\\
5.4	4.4851166532582\\
5.4	4.58866686132089\\
4.6	4.58866686132089\\
4.6	4.4851166532582\\
4.8	4.07621381886077\\
4.6	3.70986136110682\\
4.6	3.56734377850382\\
}--cycle;
\addplot [color=black, line width=2.0pt, forget plot]
  table[row sep=crcr]{%
4.8	4.07621381886077\\
5.2	4.07621381886077\\
};
\addplot [color=black, line width=2.0pt, forget plot]
  table[row sep=crcr]{%
5	2.47549019607839\\
5	3.56734377850382\\
};
\addplot [color=black, line width=2.0pt, forget plot]
  table[row sep=crcr]{%
5	4.58866686132089\\
5	5.84795321637429\\
};
\addplot [color=black, line width=2.0pt, forget plot]
  table[row sep=crcr]{%
4.9	2.47549019607839\\
5.1	2.47549019607839\\
};
\addplot [color=black, line width=2.0pt, forget plot]
  table[row sep=crcr]{%
4.9	5.84795321637429\\
5.1	5.84795321637429\\
};
\addplot [color=black, forget plot]
  table[row sep=crcr]{%
0.5	0\\
5.5	0\\
};
\draw[line width=0.2mm,dashed] (axis cs:0.5,4.8) -- (axis cs:5.5,4.8);
\end{axis}
\end{tikzpicture}%
\label{fig:stats-efreq-all-friedman}}
	}
	\caption{Box plots comparing the conditions of \ExpTwo through the dependent measures: peak hammer velocity \bvalue{\vhat}, gain \bvalue{\Ggain} and excitation frequency \bvalue{\ffrequency}.
	The lower and upper limits of the boxes represent the first and third quartiles, respectively.
	The notches represent the \SI{95}{\percent} confidence interval of the median and the whiskers show the minimum and maximum values.
	There is no significant effect of the condition on the performance as the \SI{95}{\percent} confidence intervals of the medians overlap.
	The condition marked ``Ref.'' is the \SI{4.8}{\Hz} condition from \ExpOne used as reference for \ExpTwo.
	The dashed line shows the resonance frequency.
	}
	\label{fig:stats-all-friedman}
\end{figure*}
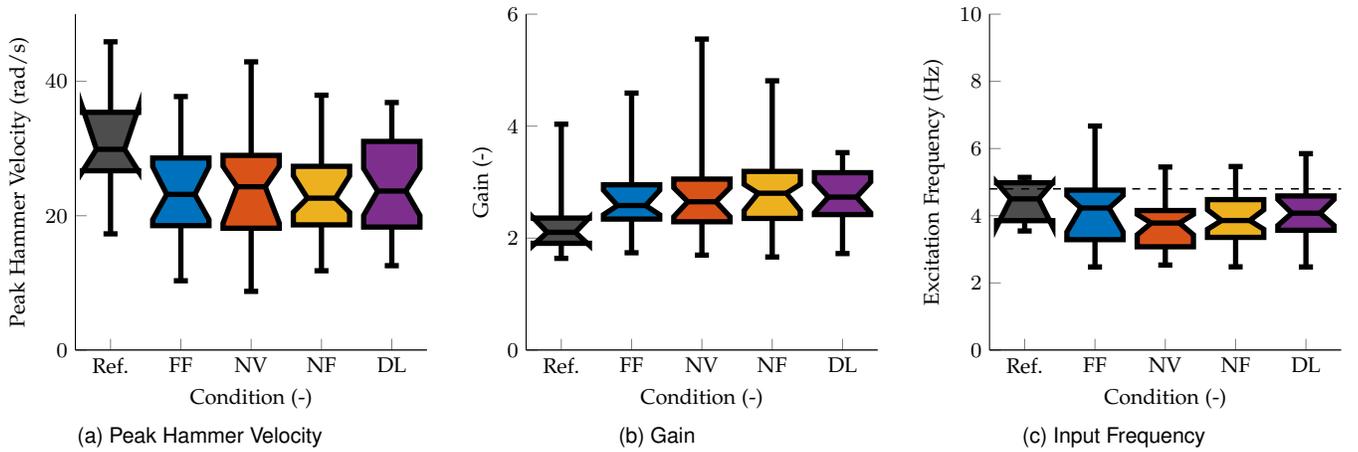

The raw measurement data are available in the archives of the 4TU Centre for Research Data \cite{Aiple2018HumanTypes}.
Fig.~\ref{fig:stats-all-friedman} shows the box plots of the dependent measures for the reference condition of \ExpOne and the four conditions of \ExpTwo: peak hammer velocity \bvalue{\vhat}, gain \bvalue{\Ggain} and excitation frequency \bvalue{\ffrequency}.

There are no effects significant at the \SI{5}{\percent} level for \bvalue{\vhat} and \bvalue{\Ggain}
(\bvalue{\vhat}:
\begin{math}\chisquarevar = \num{0.56}\end{math},
\begin{math}\pvar = \num{0.91}\end{math},
\begin{math}\nvar = \num{32}\end{math};
\bvalue{\Ggain}:
\begin{math}\chisquarevar = \num{6.86}\end{math},
\begin{math}\pvar = \num{0.076}\end{math},
\begin{math}\nvar = \num{32}\end{math}).
Friedman's test showed a significant effect at the \SI{5}{\percent} level for \bvalue{\ffrequency} (
\begin{math}\chisquarevar = \num{8.59}\end{math},
\begin{math}\pvar = \num{0.035}\end{math},
\begin{math}\nvar = \num{32}\end{math}), but Wilcoxon's test indicated no effect at the \SI{1}{\percent} level for the pairwise comparisons.
Therefore, the possible effects are weak and not consistent over the three measures.
There were also no significant effects of the order of conditions that the subjects have performed.

The estimated minimum effect sizes are
\SI{4.86}{\radian\per\s} for \bvalue{\vhat} 
(\begin{math}\sigmastdest = \SI{9.81}{\radian\per\s}\end{math},
\begin{math}n = 32\end{math}),
\num{0.31} for \bvalue{\Ggain} 
(\begin{math}\sigmastdest = \num{0.62}\end{math},
\begin{math}n = 32\end{math}),
and \SI{0.53}{\Hz} for \bvalue{\ffrequency} 
(\begin{math}\sigmastdest = \SI{1.06}{\Hz}\end{math},
\begin{math}n = 32\end{math}).

\section{Discussion}
\label{sec:discussion}
This study shows that human operators can exploit a flexible hammer to achieve higher peak velocities in a hammering task than they can with a rigid hammer.
A gain of over \SI{200}{\percent} was measured, defined as peak hammer velocity divided by peak handle velocity (e.g., \ExpOne \SI{4.8}{\Hz} condition: \SI{208}{\percent} median gain).
This is comparable to the results reported by Wolf et al. \cite{Wolf2008} for automated motion with SEAs  (\SI{272}{\percent}), but significantly lower than the results reported by Garabini et al. \cite{Garabini2011} for an analytical solution with SEAs (\SI{400}{\percent}).
The achieved velocity and gains are comparable between direct manipulation and teleoperation.
In the following, we will discuss the observed effects in more detail.

\subsection{Humans can exploit the mechanical resonance of a flexible hammer}
The results of flexible hammering in direct manipulation (\ExpOne) indicate that humans can sense and excite the mechanical resonance of a flexible hammer to achieve a higher peak velocity than with a rigid hammer (supporting hypothesis H1).
These results are coherent with research showing that humans can tune their limbs to a specific resonance frequency~\cite{Verdaasdonk2007,Hatsopoulos1996}.
An optimum working point was found for a flexible hammer with a resonance frequency of \SI{4.8}{\Hz} in terms of velocity gain and matching the resonance frequency.
Participants achieved a similar performance in terms of peak velocity and velocity gain for the condition with \SI{6.9}{\Hz} resonance frequency, but the excitation frequency was closer to the resonance frequency in the \SI{4.8}{\Hz} condition.

\subsection{Teleoperation and direct manipulation with flexible hammers offer similar task performance}
The results of hammering in teleoperated manipulation (\ExpTwo) indicate that humans can sense and excite the mechanical resonance of a flexible hammer to achieve a higher peak velocity than they can with a rigid hammer also through a teleoperation system (supporting hypothesis H2).

They also suggest that the absence of visual feedback (condition~\NVcond) does not affect the peak velocity or the velocity gain, and thus confirm sub-hypothesis H2.1.
Contrary to sub-hypothesis H2.2, the results do not show an effect of the absence of force feedback (condition~\NFcond) in the teleoperated flexible hammering task.
Sub-hypothesis H2.3 was also not confirmed by the results of \ExpTwo:
No effect of a communication delay of \SI{40}{\ms} round-trip time was observed (condition~\DLcond), although this represents a considerable communication delay compared to the execution time of one hammer strike (approximately \SI{20}{\percent} for a resonance frequency of \SI{4.8}{\Hz}).
In \ExpTwo, only one resonance frequency was studied experimentally in order to limit the parameter space.
However, based on the experience with the experiment apparatus, we believe that the results can be generalized to the full range of frequencies that were tested in \ExpOne.

\subsection{Sensitivity of the gain to the excitation frequency}
It was unexpected that no effect of visual feedback, force feedback, or communication delay could be measured.
This raises the question whether the same results could be achieved without any feedback.
As shown in Fig.~\ref{fig:resonance-velocity}, the velocity gain is greater than \num{1} for all frequencies lower than approximately \num{1.5} times the resonance frequency.
In essence, the excitation frequency does not have to match the resonance frequency exactly to obtain a higher output velocity with a flexible hammer.
Nevertheless, as we saw in \ExpOne (Fig.~\ref{fig:stats-efreq-all-exp1-friedman}), the participants changed their excitation frequency for different resonance frequencies, indicating that they tried to achieve the highest possible gain.

For higher resonance frequencies, it is difficult for a human operator to execute the motion fast enough. As shown by the results of \ExpOne, the excitation frequency was consistently below \SI{8}{\Hz} (which corresponds to the upper frequency limit of coordinated hand movements~\cite{Kunesch1989}), even though the resonance frequency of the stiffest condition was approximately \SI{10}{\Hz}.
Therefore, the highest gain can be expected for resonance frequencies between \SI{4}{\Hz} and \SI{8}{\Hz}.

\subsection{Influence of the task motion profile}
The results of \ExpTwo do not indicate an effect of the absence of force feedback on the flexible hammering task. However, for dexterous manipulation at much lower frequencies (e.g., lower than \SI{1}{\Hz}), it has been shown that force feedback has a significant effect, resulting in reduced task execution time, contact forces, and task load \cite{Sheridan1992,Richard1995,Wildenbeest2013242}.

The difference in results between dexterous manipulation experiments and our experiments might be due to a difference in the arm/hand movements executed by subjects in our hammering experiments.
Kunesch et al. differentiate between slow type~I motions at frequencies up to \SI{2}{\Hz} involving focal sensory control (e.g., tactile exploration), and fast type~II motions at frequencies between \SI{4}{\Hz} and \SI{8}{\Hz} monitored by pre-attentive sensory processes (e.g., writing, tapping, shading) \cite{Kunesch1989}.
The excitation frequencies observed in this study are in the frequency range of type~II motions and it is therefore coherent that the feedback channel has little influence on the task performance.

Research on human motor control also indicates that joint stiffness control for particular tasks is largely controlled by muscle activation patterns triggered by the cerebellum in a feedforward manner \cite{Smith1996}.
As the system dynamics did not change in \ExpTwo, it was apparently easy for the participants to build an internal model for feedforward control, independently of the type of feedback.

\subsection{Implications for the use of SEAs}
In this study, we found that it can be expected that human operators can do the flexible hammering task  in teleoperated manipulation as well as they can in direct manipulation.
This is an encouraging result for the use of SEAs in future dynamic teleoperation systems, as similar tasks, such as shaking, jolting and throwing could possibly also benefit from SEAs. This study is trailblazing a path toward the performance of real-life human dynamic tasks through teleoperation.

We see two immediate recommendations for the design of teleoperator systems with flexible tool device actuators.
1) If the quality of the force feedback in teleoperation is less critical for high-speed tasks than for slow dexterous manipulation, the force fed back to the operator could be scaled down for faster tasks without performance penalties.
Yet, scaling down the force on the handle device side of the teleoperation system reduces the power delivered by the handle device, as it is the product of force times velocity.
This could lead to a method to increase the stability of teleoperation systems without introducing high damping, as is done in passivity observer and controller approaches \cite{J.Rebelo2015}.
2) If the precise matching of excitation frequency and the resonance frequency is not critical to obtain a velocity gain, the stiffness of the flexible hammer can be driven by design requirements other than the resulting resonance frequency.
Even for flexible hammers, it is often desirable that the stiffness is sufficiently high to avoid excessive deflection.
The results of \ExpOne show that the flexible hammer could be designed with a resonance frequency \SI{40}{\percent} higher than reachable by a human operator and still achieve a velocity gain of more than \SI{200}{\percent}.
Consequently, if a higher tool stiffness is desired while still allowing human operators to achieve a good velocity gain, stiffnesses resulting in resonance frequencies up to about \SI{11}{\Hz} can be chosen.

\subsection{Teleoperation with variable stiffness actuators}
The presented findings support the notion of using SEAs as tool devices in teleoperation.
However, a teleoperation system with large oscillations as used in this study is not practical for positioning tasks.
It is therefore preferable to be able to change the mode of operation between soft and rigid according to the task at hand, as can be done with variable stiffness actuators \cite{Vanderborght2012}. 
Furthermore, Garabini et al. have shown that an additional velocity increase of \SI{30}{\percent} can be achieved by varying the stiffness during the motion \cite{Garabini2011}.
Therefore, it appears worthwhile to the authors to try to further improve the gain measured in the experiments presented in this paper by using a variable stiffness tool device actuator instead of a constant stiffness elastic actuator.

\section{Conclusion}
\label{sec:conclusion}
This paper shows that teleoperators can achieve peak velocities in a hammering task with a flexible (SEA-based) hammer of more than \SI{200}{\percent} of the peak velocities achieved with a rigid hammer.
Based on the performed experiments, we conclude that:
1)~Humans can sense and excite the mechanical resonance of a flexible hammer to achieve a higher peak velocity than they do with a rigid hammer in direct manipulation;
2)~a similar level of performance is achieved in teleoperated manipulation;
3)~the absence of visual feedback or force feedback, or a communication delay of \SI{40}{\ms} has no significant effect on the performance in terms of peak hammer velocity, gain, and excitation frequency.

\section{Acknowledgements}
\label{sec:acknowledgments}
This project was supported by the Dutch organization for scientific research (NWO) under the project grant~12161.
The authors would like to thank all participants of the experiments and the ESA Telerobotics and Haptics Lab for the support and provided facilities.

\bibliography{bibliography}
\bibliographystyle{IEEEtran}

\begin{IEEEbiography}[{\includegraphics[width=1in,height=1.25in,clip,keepaspectratio]{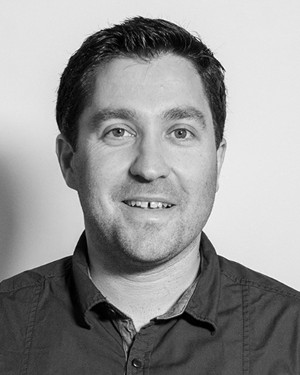}}]{Manuel Aiple}
received the MSc degree in mechanical engineering in 2010 from the University of Karlsruhe, Germany.
Since 2015, he is a PhD student at the Biomechanical Engineering department of the Delft University of Technology, Netherlands.
The topic of his PhD project is the optimization of human task performance through flexible tools in dynamic teleoperation.
\end{IEEEbiography}

\begin{IEEEbiography}[{\includegraphics[width=1in,height=1.25in,clip,keepaspectratio]{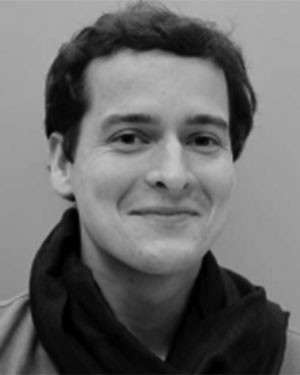}}]{Jan Smisek}
received the MSc degree in systems and control in 2011 from the Czech Technical University in Prague, Czech Republic, and the PhD degree in robotics in 2017 from the Delft University of Technology, Netherlands.
He is a robotics research engineer in the Telerobotics and Haptics Lab, with the European Space Research and Technology Centre (ESTEC) 
of the European Space Agency (ESA)
in Noordwijk, Netherlands.
His research interests include human-machine interfaces, shared autonomy, and teleoperation with long communication delays.
\end{IEEEbiography}

\begin{IEEEbiography}[{\includegraphics[width=1in,height=1.25in,clip,keepaspectratio]{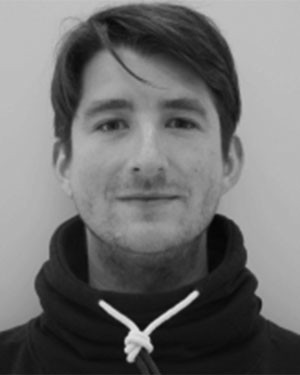}}]{André Schiele} received his Dipl. engineering degree in micro electro-mechanical systems in 2001 in Germany and the PhD degree in mechanical engineering on the topic of robotic exoskeletons in 2008 from the Delft University of Technology, Netherlands. In 2010 he founded the ESA Telerobotics \& Haptics Laboratory and led it as Principal Investigator (PI). He is PI of multiple space flight experiments conducted between Space and Earth, related with haptics and telerobotics.
He is now Associate Professor for advanced mechatronics \& telerobotics with Delft University of Technology, Faculty of Mechanical, Maritime and Materials Engineering and has published nearly one hundred papers on robotics, mechatronics, space robotics, telerobotics, and haptics.
\end{IEEEbiography}

\end{document}